\documentclass[lettersize,journal]{IEEEtran}
\usepackage{amsmath,amsfonts}
\usepackage{algorithmic}
\usepackage{algorithm}
\usepackage{array}
\usepackage{textcomp}
\usepackage{stfloats}
\usepackage{url}
\usepackage{verbatim}
\usepackage{subfloat}
\usepackage{cite}
\usepackage{gensymb}
\usepackage{color}
\usepackage{graphicx}  
\usepackage[mathscr]{eucal}
\usepackage[caption=false,font=scriptsize,labelfont=rm,textfont=rm]{subfig}
\usepackage{booktabs}
\usepackage{float}   
\usepackage{ragged2e} 
\usepackage{bbm}
\usepackage{bm}
\usepackage{xcolor}
\usepackage{hyperref}
\usepackage{multirow}
\hypersetup{hypertex=true,
	colorlinks=true,
	linkcolor=blue,
	anchorcolor=blue,
	citecolor=blue}
\usepackage{orcidlink}
\usepackage{tikz}

\begin{document}

\title{REMM:Rotation-Equivariant Framework for End-to-End Multimodal Image Matching}

\author{Han Nie\hspace{-1.0mm}$^{~\orcidlink{0000-0001-9229-6109}}$, 
Bin Luo\hspace{-1.0mm}$^{~\orcidlink{0000-0002-3040-3500}}$,~\IEEEmembership{Senior Member,~IEEE},
 Jun Liu\hspace{-1.0mm}$^{~\orcidlink{0000-0002-8943-079X}}$,
Zhitao Fu\hspace{-1.0mm}$^{~\orcidlink{0000-0002-1212-7186}}$,
 Weixing Liu\hspace{-1.0mm}$^{~\orcidlink{0000-0002-0681-5257}}$,
 and Xin Su\hspace{-1.0mm}$^{~\orcidlink{0000-0003-0957-4628}}$
 
	%

\thanks{
	This work was supported by the National Natural Science Foundation of China under Grant 41961053 and Yunnan Fundamental Research Projects under Grant 202301AT070463.	
 
	\IEEEcompsocthanksitem Han Nie, Jun Liu,  Weixing Liu and Bin Luo are with the State Key Laboratory of Information Engineering in Surveying, Mapping and Remote Sensing, Wuhan University, Wuhan 430079, China (e-mail:niehan@whu.edu.cn; luob@whu.edu.cn;liujunand@whu.edu.cn).
	\IEEEcompsocthanksitem Xin Su is with the School of Remote Sensing and Information Engineering,Wuhan University, Wuhan 430079, China (e-mail: xinsu.rs@whu.edu.cn).
	\IEEEcompsocthanksitem Zhitao Fu is with the Faculty of Land Resources Engineering, Kunming University of Science and Technology, Kunming 650031, China (e-mail:zhitaofu@kust.edu.cn).
	\IEEEcompsocthanksitem Corresponding author: Jun Liu.}


}

\maketitle

\begin{abstract}We present REMM, a rotation-equivariant framework for end-to-end multimodal image matching, which fully encodes rotational differences of descriptors in the whole matching pipeline. Previous learning-based methods mainly focus on extracting modal-invariant descriptors, while consistently ignoring the rotational invariance. In this paper, we demonstrate that our REMM is very useful for multimodal image matching, including multimodal feature learning module and cyclic shift module. We first learn modal-invariant features through the multimodal feature learning module. Then, we design the cyclic shift module to rotationally encode the descriptors, greatly improving the performance of rotation-equivariant matching, which makes them robust to any angle. To validate our method, we establish a comprehensive rotation and scale-matching benchmark for evaluating the anti-rotation performance of multimodal images, which contains a combination of multi-angle and multi-scale transformations from four publicly available datasets. Extensive experiments show that our method outperforms existing methods in benchmarking and generalizes well to independent datasets. Additionally, we conducted an in-depth analysis of the key components of the REMM to validate the improvements brought about by the cyclic shift module. Code and dataset at \url{https://github.com/HanNieWHU/REMM}.

\end{abstract}


\section{Introduction}

Image matching is widely used in remote sensing and computer vision fields such as change detection~\cite{dong2024changeclip}, visual localization~\cite{toft2020long}, and image fusion~\cite{ma2023reciprocal}. However, due to the limitations of unimodal imaging, the acquired information is often susceptible to environmental factors such as clouds, fog, and smoke. Therefore, it is necessary to comprehensively utilize multimodal data to achieve complementary advantages, improving the accuracy and reliability of subsequent applications. With the development of sensors, devices such as optical cameras, SAR, and infrared cameras have made multimodal information fusion possible. Consequently,  realizing robust multimodal image matching to effectively extract multimodal information has become an important research focus~\cite{li2023multimodal}.

In practice, geospatial data acquired via satellites can correct rotational and scale differences between multimodal images using sensor parameters. However, multimodal images acquired from UAVs or cameras cannot correct rotational and scale differences when sensor parameters are missing or inaccurate. Geometric and radiometric differences between multimodal images pose significant challenges to matching, further exacerbated by rotational and scale differences.

To address the challenge of matching multimodal images, several hand-crafted~\cite{ye2017robust}\cite{fu2018local}\cite{fan2023nonlinear}\cite{yao2022multi} and learning-based methods~\cite{chaozhen2021deep}\cite{cui2021map}\cite{zhang2024multimodal}\cite{9999700}\cite{quan2023efficient}\cite{quan2022self}\cite{quan2022deep} have been proposed to focus on the nonlinear radial difference (NRD) and geometric distortion between multimodal images. Although promising results were achieved, their experiments did not apply rotation and scaling transformations simultaneously to a pair of images. In hand-crafted traditional methods, rotation- and scale-invariant descriptors are often constructed using filtering, but these methods are mainly limited by the robustness of feature extraction compared to deep learning. In deep learning methods, input images with rotational differences lead to unpredictable feature changes, resulting in insufficient discriminative power of the descriptors due to inadequate characterization of rotational and scale features. Additionally, learning-based homologous image matching methods~\cite{dusmanu2019d2}\cite{huang2023adaptive}\cite{lee2023learning}\cite{sun2021loftr}\cite{edstedt2023dkm}\cite{edstedt2023roma} ignore the modal differences between multimodal images and it is difficult to overcome NRD and geometric distortion.

\begin{figure}[!t]
	\centering
	\includegraphics[width=0.95\linewidth]{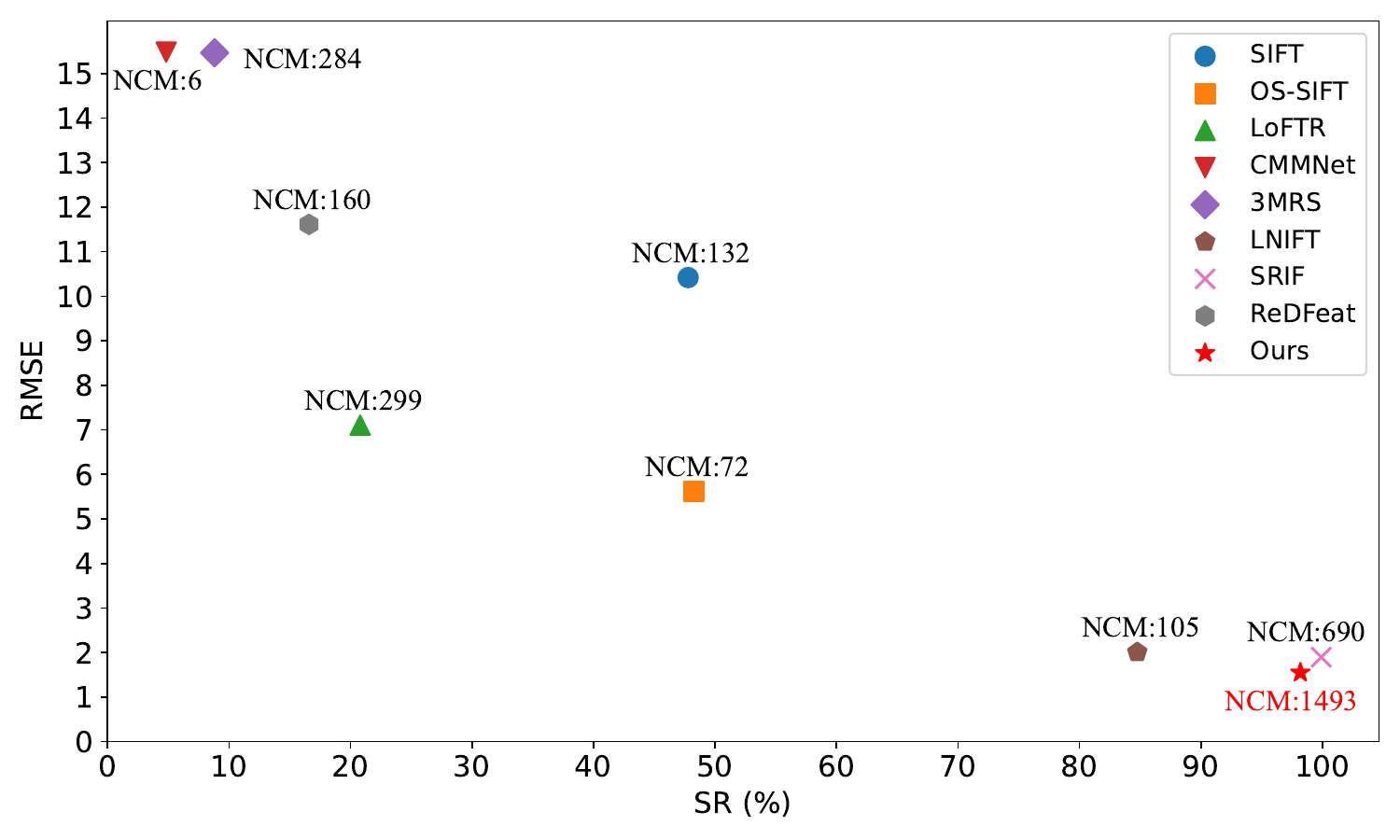} 
	\caption{On our test benchmark consisting of 33,180 pairs of images across four datasets, REMM achieved notable results. Represented by matching success rate (SR) and RMSE on the horizontal and vertical axes respectively, REMM demonstrated the second-best matching success rate, the highest NCM, and the lowest RMSE. These findings underscore the significant performance of our method. }
	\label{fig. ind}
\end{figure}

The field of multimodal image rotation and scaling matching is hampered by the current lack of multimodal image matching benchmarks that perform simultaneous rotation, scaling, and translation transformations. To remedy these shortcomings, we develop a new rotation and scaling matching benchmark for multimodal image matching that covers a wide range of sensors and includes rotation, scaling, and translation transformations. It geometrically transforms images captured by six different types of imaging platforms (SEN1, SEN2, GF-3, Google Earth, aerial, and camera) to obtain 33,180 pairs of images.

\begin{figure*}[!t]
	\centering
	\includegraphics[width=0.6\linewidth]{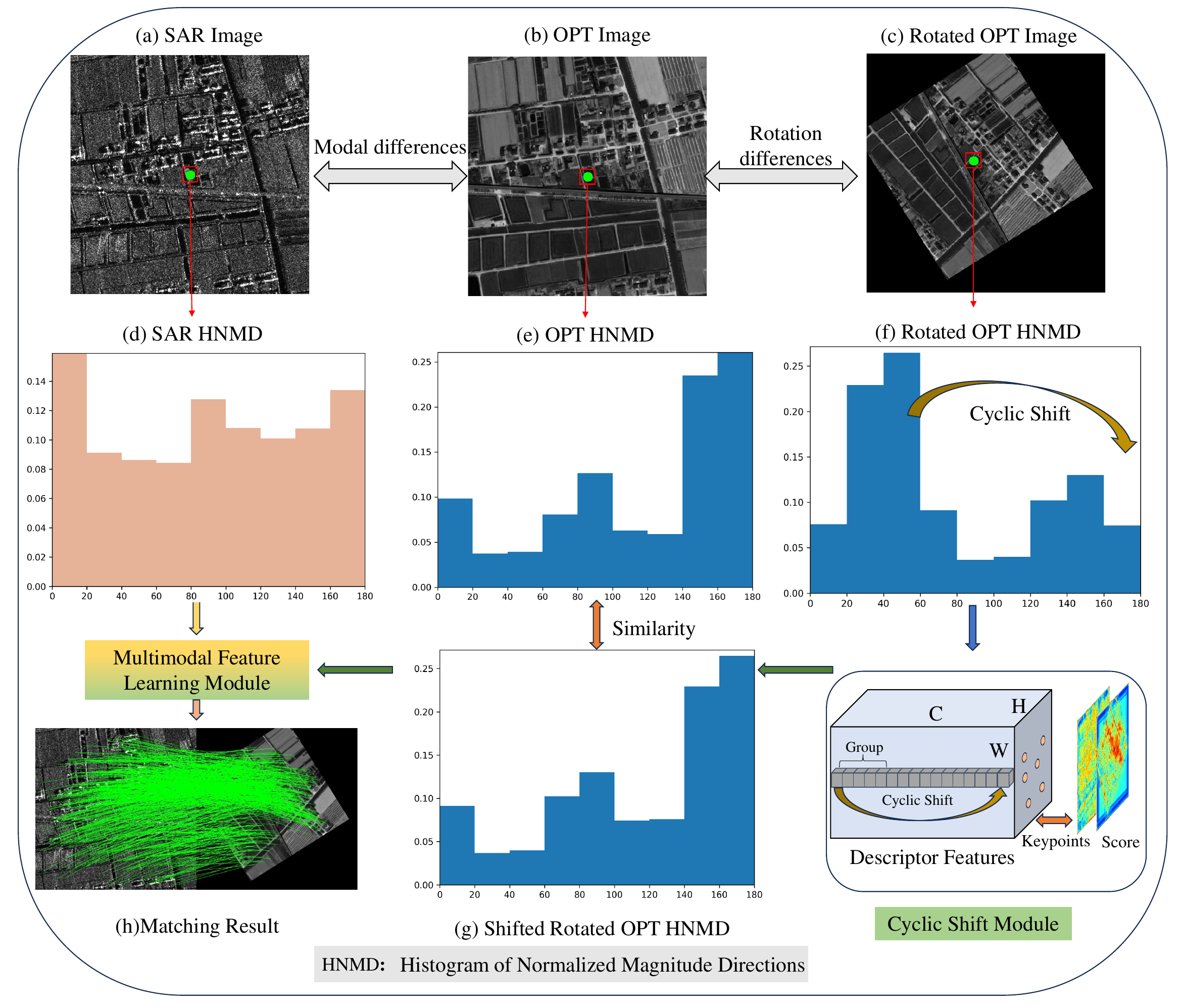}
	\caption{Our proposed REMM framework consists of multimodal feature learning module and a cyclic shift module.}
	\label{fig. remm}
\end{figure*}

Inspired by the traditional hand-crafted rotation-invariant descriptor construction method~\cite{li2018rift}\cite{fu2018local}\cite{fan2023nonlinear} and learning-based modal-invariant descriptor methods~\cite{cui2021map}\cite{zhang2024multimodal}\cite{9999700}\cite{deng2023interpretable}. We rethink the challenges of rotation-equivariant matching of multimodal images, including modal differences, rotation, and scale differences. As shown in Fig.~\ref{fig. remm}, the Histogram of Normalized Magnitude Directions (HNMD) of the unrotated multimodal images in Fig.~\ref{fig. remm} (a) and (b) exhibits large modal differences, and existing learning-based matching methods can achieve better results. However, when matching Fig.~\ref{fig. remm} (a) and (c), the results are much less effective, mainly due to the difficulty of encoding rotational differences in existing methods. Therefore, Quan et al. developed a coarse-to-fine matching method using two network models, one for correcting rotational transformations and the other for extracting modal invariant features~\cite{quan2023novel}. However, the complexity of the proposed matching pipeline is high, making real-time matching difficult. Therefore, in this paper, we design a cyclic shift module to rotationally encode Fig.~\ref{fig. remm} (f) to obtain Fig.~\ref{fig. remm} (g) which has high similarity with Fig.~\ref{fig. remm} (e) but exhibits large modal differences with Fig.~\ref{fig. remm} (d), making matching difficult. Finally, we learn modal-invariant features using the multimodal feature learning module to achieve high-performance rotation-equivariant matching, as shown in Fig.~\ref{fig. ind}. 

In summary, We propose a rotation-equivariant framework for end-to-end multimodal image matching (REMM). It includes multimodal feature learning module and cyclic shift module.
The contributions of this paper can be summarized as follows:
\begin{itemize}
	\item
    To break through the bottleneck of existing learning-based matching methods in achieving multimodal image rotation-equivariant matching, we propose a rotation-equivariant framework for end-to-end multimodal image matching (REMM), which consists of a multimodal feature learning module and a cyclic shift module.
		
\item
    We propose an innovative cyclic shift module suitable for constructing rotation-equivariant descriptors for multimodal images, ensuring reliable texture matching over a $360\degree$ rotation range.
		
\item
    We curated a new rotation and scaling matching benchmark for multimodal image matching, which performs rotation, scaling and translation transformations on multiple sensor data.
\item
    After rigorous experiments, we show that our method outperforms other strong baselines on four datasets, achieving the state-of-the-art (SOTA) performance in multimodal rotation-equivariant matching.
\end{itemize}

The rest of this paper is organized as follows. Section \ref{2} describes the related works on multimodal image matching. Section \ref{3} introduces our proposed REMM. Section \ref{4} presents the experimental results and analysis. Section \ref{5} gives the discusses of this paper. Section \ref{6} presents the paper conclusions.

\section{Related Work}
\label{2}
In this section, previous studies are presented by categorizing them into two groups: traditional handcrafted methods and learning-based methods. Where traditional handcrafted methods are categorized into area- and feature-based methods. The learning-based methods are categorized into Detector-based and Detector-free methods.
\subsection{Traditional Handcrafted Methods}
{\bf{Area-based methods:}} Area-based matching primarily calculates the similarity between keypoint image blocks directly on pixel values and identifies image blocks with high similarity as matching point pairs. Similarity methods include the sum of squared differences (SSD), normalized cross correlation (NCC), mean absolute difference (MAD), and mutual information (MI)~\cite{suri2009mutual}. These methods are computationally slow, prompting experts and scholars to reduce computational cost and improve efficiency using the fast fourier transform~\cite{wan2016illumination}. However, this approach is sensitive to differences in multimodal images, making matching particularly challenging, especially due to the speckle noise in SAR images. To mitigate the effect of imaging differences in multimodal images on matching, researchers introduced phase correlation (PC) to counteract NRD in the frequency domain. Furthermore, the Fourier Transform within PC can partially address translation, scaling, and rotation differences in multimodal images~\cite{reddy1996fft}; however, it remains highly sensitive to substantial rotation and scale variations. Some of the most representative methods include the HOG~\cite{li2013matching}, HOPC~\cite{ye2017robust}, CFOG~\cite{ye2019fast} and SFOC~\cite{ye2022robust} methods. Both HOG and HOPC methods are sparsely sampled feature representations, which often result in ignoring important structural information, thereby reducing the robustness of matching. Conversely, CFOG and SFOC map the image into a high-dimensional fine feature representation on a pixel-by-pixel basis, thereby obtaining more discriminative features and significantly improving matching accuracy. Regarding the efficiency and robustness of comprehensive alignment, dense feature-based methods are advantageous for handling differences in multimodal images; however, their lack of scale and rotational invariance can limit practical applications.

{\bf{Feature-based methods:}} The Area-based methods mentioned above struggles to handle rotational and scale transformations between multimodal images. Although geospatial data can reduce rotational and scale differences between multimodal images using sensor parameters, if the sensor parameters are missing, the image cannot be corrected, and matching cannot be performed~\cite{zhu2023advances}. Feature-based matching methods often do not require any sensor parameters. These methods rely on saliency features between images, achieving pixel-level correspondence through the steps of keypoint detection, main orientation estimation, and description. Typical approaches include SIFT~\cite{lowe2004distinctive} and SURF~\cite{bay2008speeded}, which are constructed based on gradient features. SIFT and SURF methods can find correspondences between homologous image pairs with scale and rotation differences. However, matching becomes challenging when nonlinear radiance differences between multimodal images cause gradient instability. In order to achieve robust matching between multimodal images, a lot of research has been carried out on the challenges of multimodal image rotation, scale difference and NRD. In general, these methods focus on extracting multi-scale and multi-directional filtering alternatives to gradient features in homologous methods to obtain more robust descriptors~\cite{fu2018local}~\cite{fan2023nonlinear}~\cite{yao2022multi}. RIFT~\cite{li2018rift} algorithm computes the specialty index map (maximum index map) based on the phase consistency algorithm, achieving rotation invariance for the MIM map. To further enhance matching accuracy and efficiency, LNIFT~\cite{li2022lnift} uses a local normalization filter to reduce differences between multimodal images, subsequently describing them, which significantly improves the matching accuracy. Although these algorithms can overcome the NRD of multimodal images well, they are still sensitive to scale and rotation variations.

\subsection{Learning-based Methods}
{\bf{Detector-based methods:}} With the advancement of deep learning, these methods show great potential compared to traditional approaches. Further research is needed to study the matching of multimodal images involving rotation and scale. The data-driven approach based on LIFT~\cite{yi2016lift} achieves better results, which indicates that the descriptors extracted using CNN are more robust and discriminative. Current mainstream detector-based methods can be categorized into two groups: \textit{(1)} In the traditional matching pipeline, deep learning techniques serve as preprocessed modal translation networks or replace the keypoint detection or description to create a new matching pipeline. \textit{(2)} Joint keypoint detection and description, constructing an end-to-end matching pipeline.

In the first category, to reduce significant NRD between multimodal images, Generative Adversarial Networks (GANs) are introduced to transform images to the same modality, effectively improving matching performance but increasing computational cost~\cite{nie2022dual}~\cite{merkle2018exploring}. Experts and scholars are also inspired by keypoints detection algorithms, such as SuperPoint~\cite{detone2018superpoint} for self-supervised learning of keypoint detection, and Key.Net~\cite{barroso2019key}, which designs a new keypoint detection loss. Zhang et al.~\cite{zhang2023multilevel} proposed a new multimodal image keypoint detection algorithm, yielding keypoints that are more favorable for image matching compared to traditional methods. Many experts and scholars have constructed feature learning networks by extracting image blocks based on keypoints to improve the accuracy of matching~\cite{quan2023efficient}~\cite{quan2022self}~\cite{quan2022deep}. For example, SDNet~\cite{quan2022self} employs a self-distillation feature learning method to extract robust features from image blocks.

A clear limitation exists in the first category of research combining traditional matching pipelines with deep learning methods, as the latter cannot interact with the steps in the traditional pipeline to jointly achieve more robust matching results. In the second category, to further enhance the mutual reinforcement of the modules in the pipeline, many experts and scholars propose end-to-end matching pipelines~\cite{tian2019sosnet}~\cite{dusmanu2019d2}. D2Net~\cite{dusmanu2019d2} optimizes keypoint detection and descriptors simultaneously, achieving better results. The aforementioned end-to-end D2Net method with non-learnable non-maximum suppression layers degrades performance on multimodal images. CMMNet~\cite{chaozhen2021deep} adapts the D2Net method for multimodal images. MAP-Net~\cite{cui2021map} constructs an end-to-end method for multimodal images using spatial pyramid aggregation pooling (SPAP) and an attention mechanism. MICM~\cite{zhang2024multimodal} extracts invariant features using deep convolutional neural networks and the transformer attention mechanism, achieving superior results. ReDFeat~\cite{9999700} constructs an end-to-end matching method coupling keypoint detection and description, introduces learnable non-maximum suppression layers, and achieves superior results on small rotation transformations of multimodal images. All the aforementioned methods struggle to support multimodal image matching with large rotations, making it particularly important to achieve rotational invariance of descriptors for robust matching of such images. Inspired by ~\cite{cui2021map}\cite{zhang2024multimodal}\cite{9999700}, we propose an end-to-end rotation-equivariant matching framework to achieve robust matching of large rotated multimodal images.

{\bf{Detector-free methods:}}
Although detector-based methods can reduce the search space, they often face challenges with viewpoint transformations and weakly textured regions. To address this issue, detector-free matching methods remove feature detectors, achieving dense matching results and significantly enhancing matching performance. The main detectorless methods include CNN-based~\cite{truong2020glu} and transformer-based~\cite{sun2021loftr} matching techniques. Transformer-based matching methods outperform CNN-based methods due to their global receptive field and long-range dependencies.

LoFTR~\cite{sun2021loftr} introduced the use of transformers to achieve dense matching results in weakly textured regions, significantly enhancing matching performance. DKM~\cite{edstedt2023dkm} designs dense feature matching networks for addressing multi-viewpoint matching challenges. RoMa~\cite{edstedt2023roma} integrates a Markov chain to enhance the matching process and utilizes robust regression loss to improve accuracy, achieving notable advancements. However, these methods struggle with multimodal images. XoFTR~\cite{tuzcuouglu2024xoftr} addresses multimodal differences by incorporating masked image modeling pre-training and fine-tuning with pseudo-thermal image augmentation, achieving excellent performance on RGB-TIR images. However, they still face challenges in matching large rotated images.


\section{Method}
\label{3}
\begin{figure*}[!t]
	\centering
	\includegraphics[width=0.7\linewidth]{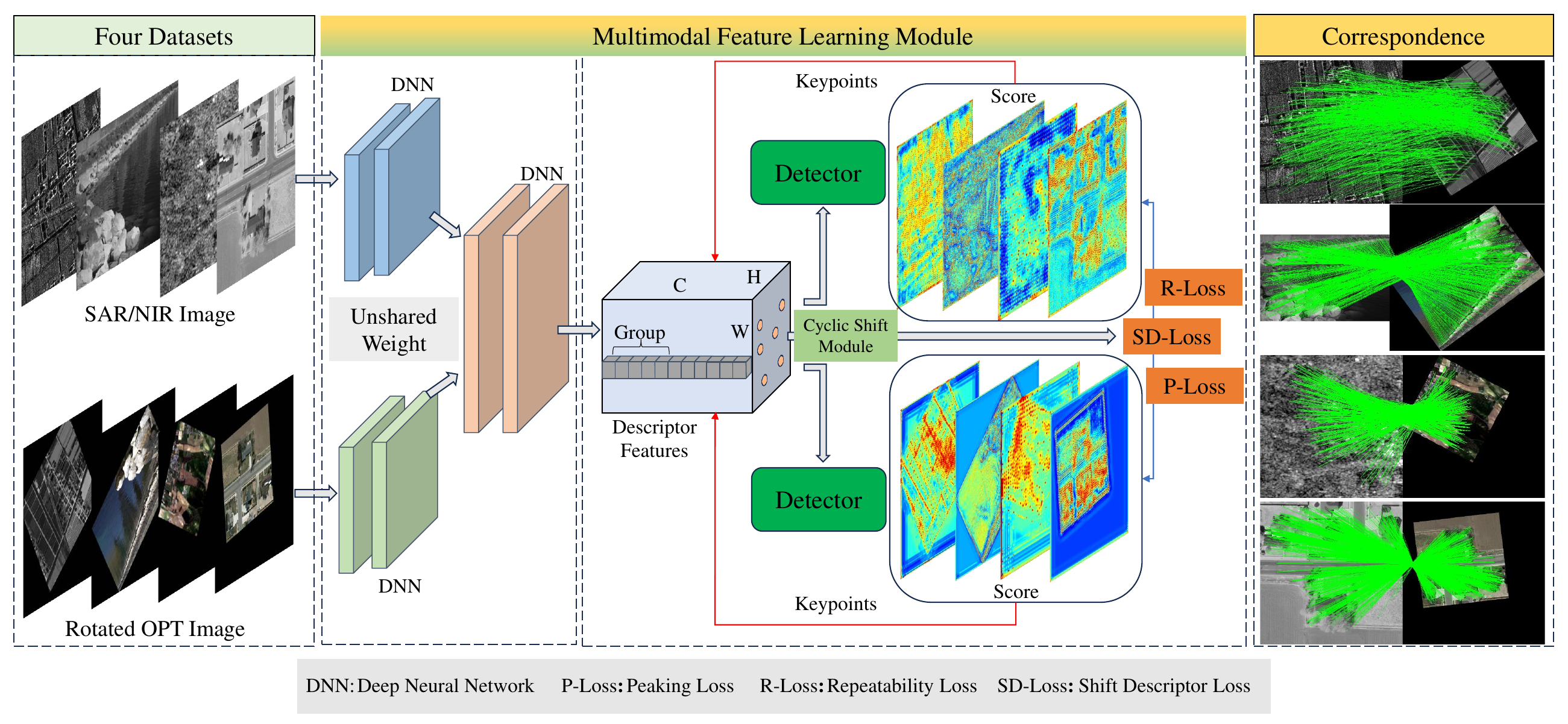}
	\caption{Our proposed REMM framework consists of multimodal feature learning module and a cyclic shift module.}
	\label{fig. 3mm}
\end{figure*}

In this paper, we propose a rotation-equivariant framework for end-to-end multimodal image matching (REMM), as shown in Fig.~\ref{fig. 3mm}. This section describes the overall framework consisting of two main modules. Each module and loss function are described in detail.

\subsection{Problem Formulation}
We develop a rotation-equivariant framework for end-to-end multimodal image matching, called REMM. The multimodal images with rotational and scale differences are denoted as $X$ and $Y$. Our aim is to match $X$ with the reference image $Y$ to determine the transformation matrix $H$, as shown in Fig.~\ref{fig. 3mm}. Unifying multimodal feature learning and descriptor rotation coding within a single framework allows the two modules to optimize each other effectively. The proposed method consists of two main parts: 

(1) Multimodal feature learning module that aims to extract common features among multimodal images. The non-aligned features of multimodal images are first filtered by Deep Neural Network (DNN) layer without shared weights. Then, the aligned features of multimodal images are extracted by DNN layer with shared weights. On the basis of aligned features, keypoints are obtained by detector. 

(2) After extracting the descriptors from the keypoint detection results, the cyclic shift module is used to rotationally encode the descriptors, obtaining rotation-equivariant descriptors.

\subsection{Multimodal feature learning module}
Due to the complex modal differences present in multimodal images, the key to extracting modal-invariant descriptors is to distinguish between non-aligned and aligned features. Inspired by multimodal feature extraction networks\cite{9999700}\cite{deng2023interpretable}, we introduce a multimodal feature learning module to obtain aligned features.

\begin{figure}[!t]
	\centering
	\includegraphics[width=0.8\linewidth]{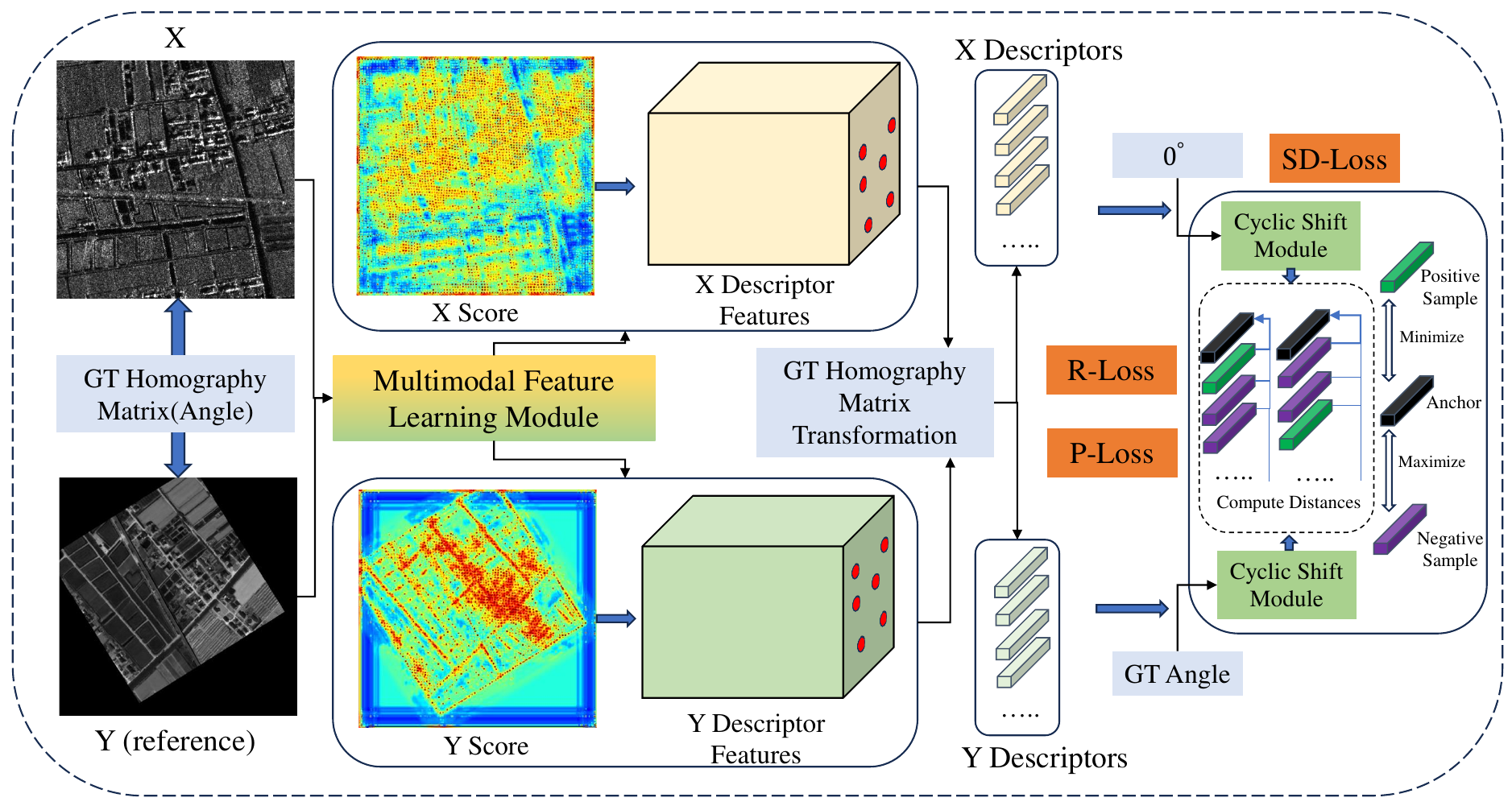}
	\caption{The training phases of our method REMM.}
	\label{fig. 4train}
\end{figure}

In order to map the multimodal features to the same feature space, as shown in Fig.~\ref{fig. 3mm}, we first filter the non-aligned features $ NF_X $ and \( NF_Y \) of the multimodal image \( X \) and \( Y \) using DNN layer without shared weights, and then extract the aligned features \( PF_X \) and \( PF_Y \) using DNN layer with shared weights. The Ground Truth (GT) homography matrix and rotation angle between are \( GT_H \) and \( GT_\theta \). In the training phase, we obtain the aligned features, i.e., the descriptor features \( D_X \) and \( D_Y \). We then obtain the 
keypoint score map \( S_X \) and \( S_Y \) through the detector. We spatially transform \( D_Y \) and \( S_Y \) based on \( GT_H \) to correct the geometric transformation differences in feature maps to obtain $CD_Y$ and $CS_Y$, as shown in Fig.~\ref{fig. 4train}. This process can be described as:
\begin{equation}
\begin{split}
D_X &= DNN_{shared}(DNN_{unshaerd}(X))
\end{split}
\label{eq:1}
\end{equation}
\begin{equation}
\begin{split}
D_Y &= DNN_{shared}(DNN_{unshaerd}(Y))
\end{split}
\label{eq:2}
\end{equation}
\begin{equation}
\begin{split}
S_X &= Detector(D_X)
\end{split}
\label{eq:3}
\end{equation}
\begin{equation}
\begin{split}
S_Y &= Detector(D_Y)
\end{split}
\label{eq:4}
\end{equation}
\begin{equation}
\begin{split}
grid_Y= F_{Affgrid}(GT_H)
\end{split}
\label{eq:5}
\end{equation}
\begin{equation}
\begin{split}
CD_Y= F_{Affsample}(D_Y, F_{Affgrid}(GT_H))\\
CS_Y= F_{Affsample}(S_Y, F_{Affgrid}(GT_H))\\
\end{split}
\label{eq:6}
\end{equation}
where grid generator ($F_{Affgrid}$) and sampler ($F_{Affsample}$) from spatial transformer network (STN)~\cite{jaderberg2015spatial}.

In addition, recent studies have found that Non-Maximum-Suppression (NMS) in key point detection is non-differentiable, causing the gradient to vanish, which makes it difficult to capture subtle changes in local scores. Therefore, many differentiable keypoint detectors have been proposed, showing strong performance in keypoint detection~\cite{9726928}\cite{9999700}. Our multimodal feature learning module also incorporates a differentiable keypoint detector.

\begin{figure}[!t]
	\centering
	\includegraphics[width=0.8\linewidth]{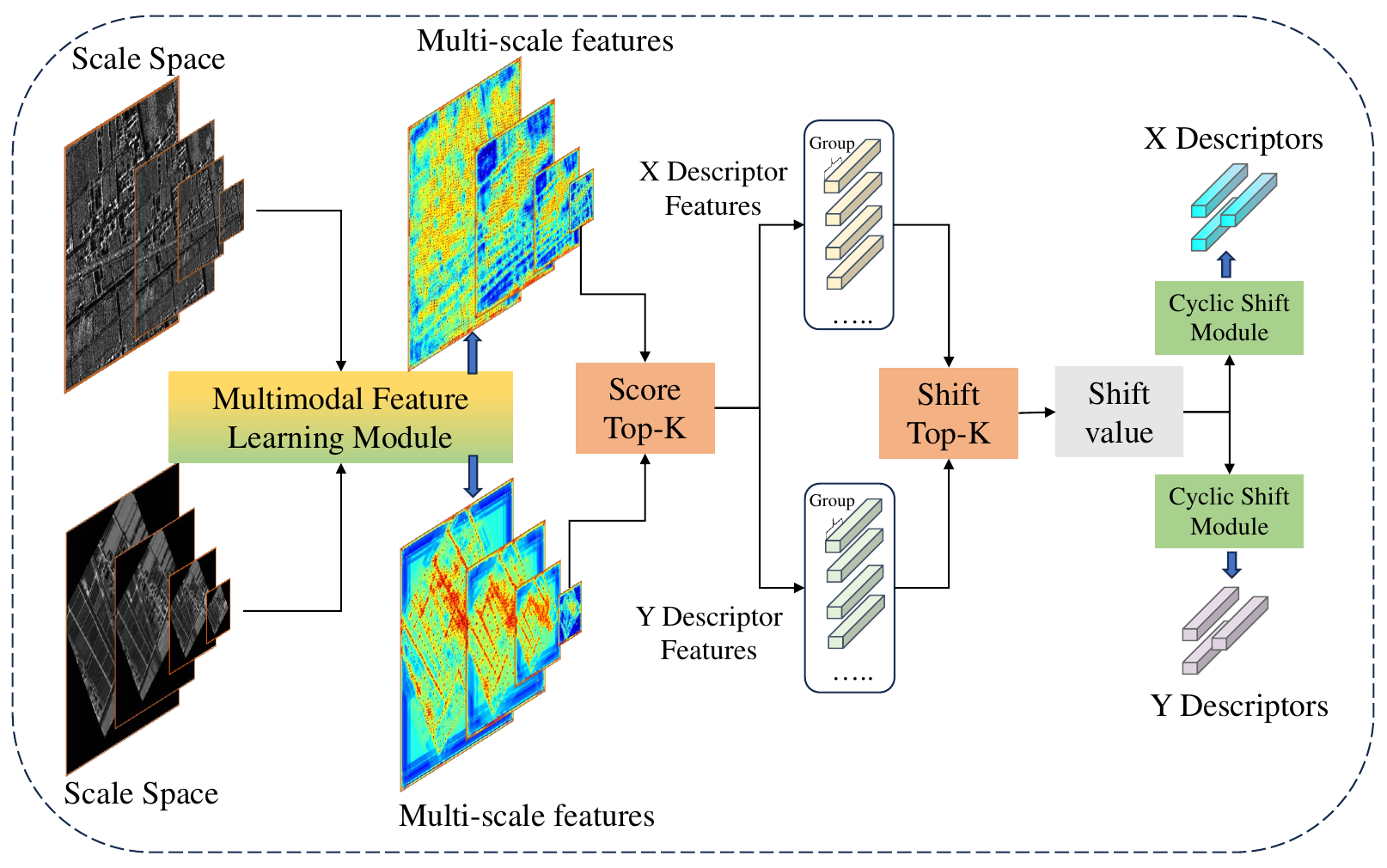}
	\caption{The testing phases of our method REMM.}
	\label{fig. 5test}
\end{figure}

Although the cyclic shift module was designed to rotationally encode the descriptors during the training phase, it does not ensure robustness to scale transformations. Therefore, we constructed a multi-scale space in the testing phase, extracted keypoints and descriptors for each scaled image, and finally selected the Top-k points based on the score as the final keypoints, as shown in Fig.~\ref{fig. 5test}.

\subsection{Cyclic shift module}
Despite our implementation of modal invariant feature extraction through the multimodal feature learning module, it is still difficult to extract rotation-invariant descriptors, as shown in Fig.~\ref{fig. remm} (e) and (f). The key issue is that the descriptors obtained from the input data with rotational differences are not rotationally invariant in the spatial dimension, and therefore equivariant matching results cannot be achieved.

\begin{figure}[!t]
	\centering
	\includegraphics[width=0.9\linewidth]{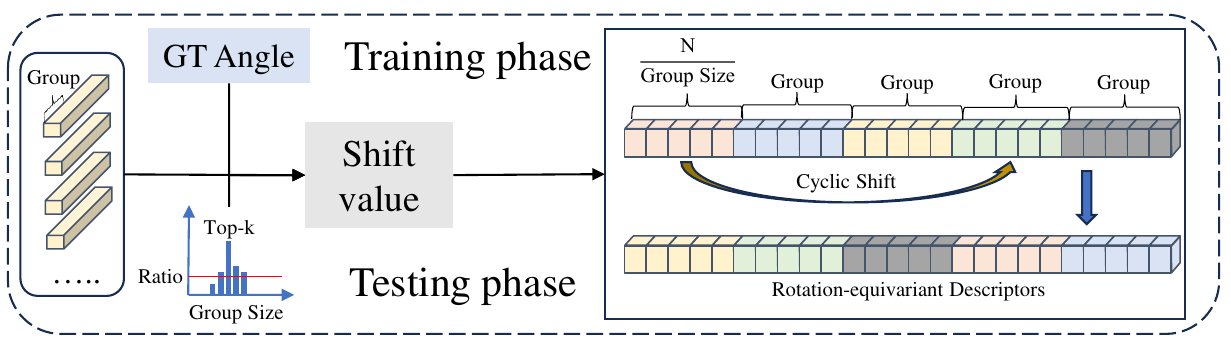}
	\caption{The training and testing phases of cyclic shift module.}
	\label{fig. 6cf}
\end{figure}

As described in the introduction, rotational differences in multimodal images can lead to differences in the spatial dimensions of the descriptors. Our approach introduces an innovative cyclic shift module designed for constructing rotation-equivariant descriptors for multimodal images, as show in Fig.~\ref{fig. 6cf}. In the training phase, the true rotation angle difference between images, denoted as $GT_\theta$, is known. After obtaining the descriptors $D_X$ and $D_Y$, we stretch these descriptors according to the group size $G_{size}$ to form grouped descriptors, similar to Fig.~\ref{fig. remm} (e) and (f). The value of cyclic shift $YS_{value}$ is calculated based on the true rotation angle difference $GT_\theta$. Subsequently, we obtain the rotation-equivariant descriptors $RD_X$ and $RD_Y$ using the cyclic shift module. This process can be described as:
\begin{equation}
\begin{split}
RD_X= Shift(D_X, 0)
\end{split}
\label{eq:7}
\end{equation}
\begin{equation}
\begin{split}
YS_{value}= (\frac{GT_\theta}{360}* G_{size}) \bmod (G_{size})
\end{split}
\label{eq:8}
\end{equation}
\begin{equation}
\begin{split}
RD_Y= Shift(D_Y, YS_{value})
\end{split}
\label{eq:9}
\end{equation}
where $shift$ denotes cyclic shift operation and $mod$ denotes modulo operation.


In the testing phase, we first take the descriptors $D_X$ and $D_Y$ obtained from the multimodal feature learning module and stretch them according to the group size $G_{size}$ to obtain the grouped descriptors. Unlike the training phase, we do not know the true rotation angle $GT_\theta$. Therefore, we extract the first channel of the stretched group descriptor as the orientation mapping histogram. We predict the Top-k to determine the cyclic shift values $XS_{value}$ and $YS_{value}$, and then perform multiple cyclic shifts for each descriptor to obtain rotation-equivariant descriptors, as show in Fig.~\ref{fig. 6cf}. This process can be described as:
\begin{equation}
\begin{split}
XS_{value}= TopK(D_X)
\end{split}
\label{eq:10}
\end{equation}
\begin{equation}
\begin{split}
YS_{value}= TopK(D_Y)
\end{split}
\label{eq:11}
\end{equation}
\begin{equation}
\begin{split}
RD_X= Shift(D_X, XS_{value})
\end{split}
\label{eq:12}
\end{equation}
\begin{equation}
\begin{split}
RD_Y= Shift(D_Y, YS_{value})
\end{split}
\label{eq:13}
\end{equation}
where $Topk$ denotes the extraction of cyclic shift values in the first dimension of the descriptor.
\subsection{Loss function}
We train our REMM by constructing training data using random rotations and scale transformations on the aligned images. Since REMM consists of multimodal feature learning module and cyclic shift module, the main purpose of the multimodal feature learning module is to extract modality invariant keypoints and descriptors in multimodal images. The main purpose of our cyclic shift module is to rotationally encode the modal invariant descriptors to obtain rotation-equivariant descriptors. Therefore, our loss function includes keypoint detection loss and shift descriptor loss.
\subsubsection{Keypoint detection loss}
There are significant geometric deformations and NRDs between multimodal images, and extracting robust keypoint detection results is a prerequisite for achieving equivariant matching. In this paper, we focus on the construction of rotation-equivariant descriptors. Similar to ReDfeat~\cite{9999700} and ALIKE~\cite{9726928}, we train our keypoint detection using their proposed peaking loss $L_P$ and repeatability loss $L_R$.

\subsubsection{Shift descriptor loss}
After passing through the multimodal feature learning module to get the modal invariant descriptors, we obtain $D_X$ and $D_Y$. Later, we acquire the rotation-equivariant descriptors $RD_X$ and $RD_Y$ after passing through the cyclic shift module as described in Eq.~\eqref{eq:7}~\eqref{eq:8}~\eqref{eq:9}.
\begin{equation}
L_{SD}(RD_X,RD_Y) = \sum_{i}^{n} -log\frac{exp(sim(d_i^1,d_i^2)/\tau)}{\sum_{k=1}^{n}exp(sim(d_i^1,d_i^2)/\tau)}
\label{eq:sd}
\end{equation}
where $i$ denotes any descriptor within the batch and $N$ denotes the number of all descriptors within the batch. $sim$ denotes cosine similarity, $\tau$ denotes softmax temperature.

Overall, our final loss function contains three parts defined as:
\begin{equation}
L_{all} = \lambda_1 L_{P} + \lambda_2 L_{R} + \lambda_3 L_{SD}
\label{eq:ALL}
\end{equation}
where $\lambda_1$, $\lambda_2$ and $\lambda_3$ are trade-off hyper-parameters.

\section{Experiments}
\label{4}
In this section, we present the experimental setup, data, and results. Detailed descriptions of our implementation and the four datasets used are provided in section~\ref{Datasets}. Due to the absence of rotation-invariant and scale-invariant benchmarks for multimodal image matching, Section~\ref{Benchmark} introduces our proposed benchmarks. Section~\ref{result} details the comparison experiments between our REMM method and baseline algorithms. Our REMM was compared with eight baseline or state-of-the-art algorithms: SIFT~\cite{lowe2004distinctive}, OS-SIFT~\cite{xiang2018sift}, LoFTR~\cite{sun2021loftr}, CMMNet~\cite{chaozhen2021deep}, 3MRS~\cite{fan20223mrs}, LNIFT~\cite{li2022lnift}, SRIF~\cite{li2023multimodal} and ReDFeat~\cite{9999700}. All implementations of the comparison methods used in the experiments were provided by the authors or obtained from publicly available software. Section~\ref{xr} presents ablation experiments aimed at evaluating the key components of REMM. 

\subsection{Implementation details}
\label{indetails}
Here, according to the test dataset setup~\ref{teet_data_setup}, we collected 316 multimodal image pairs from four datasets, which were expanded to obtain 33,180 test image pairs for a comprehensive evaluation of our proposed REMM method.  Experiments were conducted using formal implementations of each method. For a fair comparison, we used the same test dataset and set the maximum number of extracted keypoints to 5000, with the exception of LoFTR and 3MRS. LoFTR is a detector-free matching method, faces challenges in adjusting the maximum number of extracted keypoints. 3MRS only provide binary code, which complicates modification. The network ran on NVIDIA GTX 4090 with batch size of 4. We employed the Adam optimizer for network optimization, gradually decreasing the learning rate over epochs, with the final rate dropping to zero.

\subsection{Datasets}
\label{Datasets}
\subsubsection{OPT-SAR}
To verify the superiority of our proposed method, we train and test it on two different types of optical and SAR image datasets.

\textit{(a)} Sentinel-1 and Sentinel-2 (SEN1-2) dataset~\cite{schmitt2018sen1} is the first OPT-SAR dataset we used. The dataset contains 282384 optical and SAR image blocks of size 256$\times$256 consisting of multiple seasons from all over the world. These images contain a variety of land covers such as mountains, towns, rivers, farmlands, etc., from which we randomly select 2000 pairs of optical and SAR images for training and 100 pairs for testing. The constructed training and test sets are subsets of the SEN1-2 dataset. Compared with other SAR images, the SAR images obtained by Sentinel-1 satellite have more obvious noise. As shown in Fig.~\ref{datasets_img} (a).

\textit{(b)} The second OPT-SAR dataset~\cite{9204802} consists of SAR images obtained by Chinese GaoFen-3 (GF-3), a multipolarized C-band SAR satellite, and optical images of the corresponding areas collected from Google Earth, with a resolution of 1 meter. The dataset contains 2673 pairs of optical and synthetic aperture radar (SAR) image blocks with a size of 512$\times$512, and its collection area includes China, France, America and India. From these, we selected 2011 pairs of optical and SAR images for training and 100 pairs for testing. As shown in Fig.~\ref{datasets_img} (b).

\subsubsection{OPT-NIR}
To verify the superiority of our proposed method, we train and test it on two different types of optical and NIR image datasets.

\begin{figure}[h]
	\centering
	\includegraphics[width=1\linewidth]{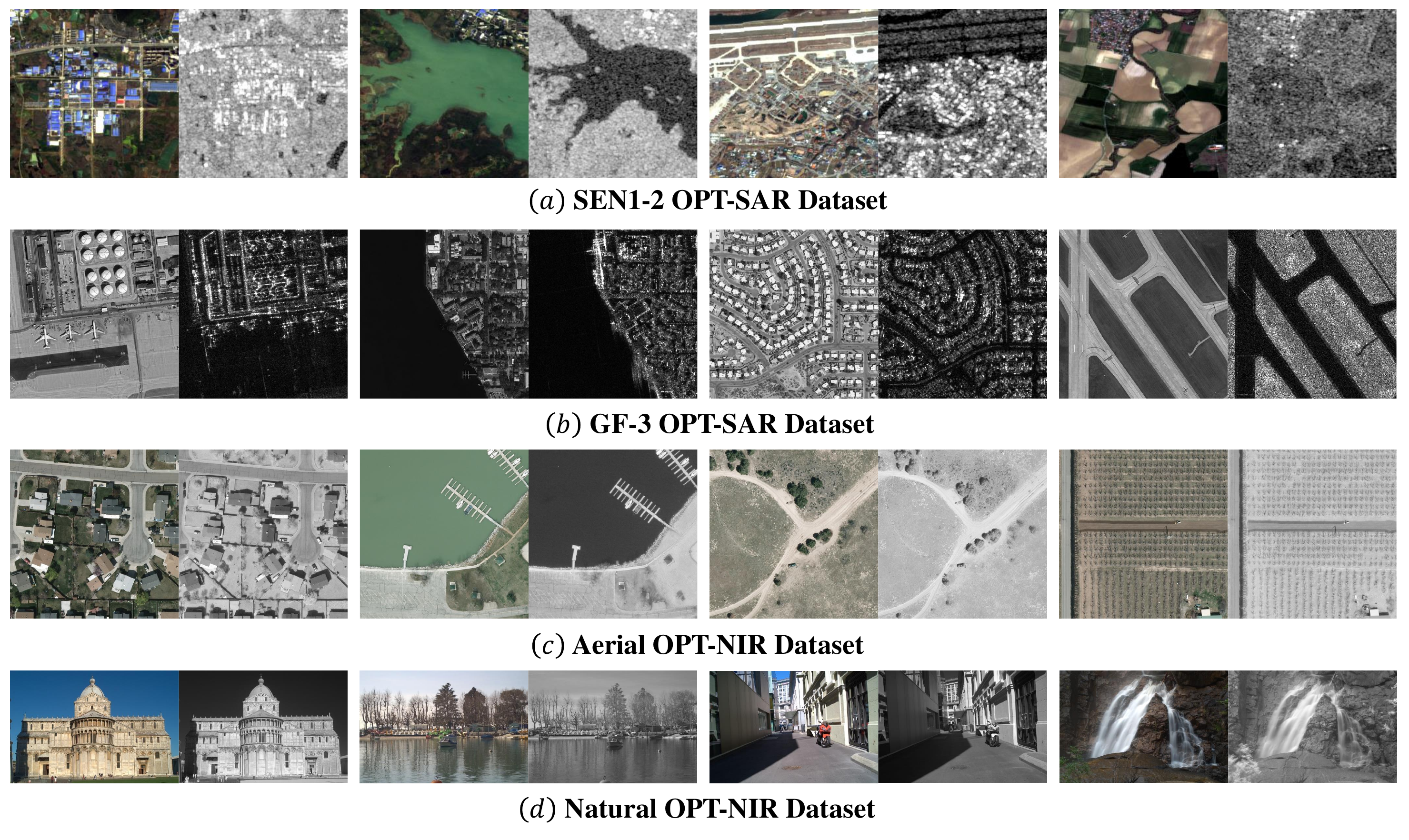}
	\caption{Some examples of the four datasets.}
	\label{datasets_img}
\end{figure}

\textit{(c)} Vehicle Detection in Aerial Imagery (VEDAI) dataset ~\cite{razakarivony2016vehicle} is the
first OPT-NIR dataset we used. This dataset is a dataset for evaluating small vehicle detection in aerial images. It contains more than 1200 pairs of 512$\times$512 optical and NIR images. From these we randomly select 1100 pairs of optical and NIR images for training and 100 pairs of images for testing. As shown in Fig.~\ref{datasets_img} (c).

\textit{(d)} The second OPT-NIR dataset~\cite{brown2011multi} contains a composition of optical and near-infrared (NIR) images of various scenes such as countryside, buildings, streets and cities. This dataset contains 477 pairs of optical and NIR image compositions with an average size of 983$\times$686. From these we randomly select 420 pairs of optical and NIR images for training and 16 pairs for testing.  As shown in Fig.~\ref{datasets_img} (d).

\begin{table*}[!h]
	\caption{Basic information about our benchmarks.
		\label{benchmarks}}
	\centering
	\renewcommand{\arraystretch}{1.3}
	\footnotesize
	\scalebox{0.8}{
		\begin{tabular}{ccccccc}
			\toprule[2pt] 
Datasets     &OPT Source &SAR/NIR Source & Size   &Train/Test Number  &Expanded Test Number &Character \\\hline
\textit{(a)} OPT-SAR~\cite{schmitt2018sen1} &SEN2  &SEN1  &256$\times$256 &2000/100   &10500  & Satellite remote sensing    \\
\textit{(b)} OPT-SAR~\cite{9204802} &Google Earth &GF-3  &512$\times$512  &2011/100   &10500    &Satellite remote sensing     \\
\textit{(c)} OPT-NIR~\cite{razakarivony2016vehicle} &Aerial image &Aerial image  &512$\times$512  &1100/100   &10500     &Aerial image       \\
\textit{(d)} OPT-NIR~\cite{brown2011multi} &Camera &Camera &983$\times$686  &420/16   &1680     & Multiple scenes image     \\
\hline
            \toprule[1.0pt]
	    \end{tabular}}
\end{table*}

\subsection{Benchmark}
\label{Benchmark}
Rotation- and scale-invariant matching of multimodal images has received increasing attention, and the lack of rotation- and scale-invariant benchmarks has led to slow development of feature-based research. In this paper, we establish a test benchmark for rotation and scale invariant matching of multimodal images. We show the basic information of our constructed benchmarks in Table~\ref{benchmarks}.

\subsubsection{Test Datasets Setup}
\label{teet_data_setup}
In order to evaluate the robustness of the multimodal image rotation and scale matching methods, we randomly selected a certain number of optical and SAR images, and optical and near-infrared images to get the data subsets in the four datasets, respectively. First, we randomly rotated each pair of multimodal images at {10\degree} intervals to obtain a rotation angle covering a {360\degree} frame. Meanwhile, three scales are obtained by randomly selecting scales among $[0.5,0.8]$, $[0.8,1]$ and $[1,1]$, where $[1,1]$ means no scale transformation. After combining the scales and angles, each multimodal image pair is expanded to obtain 105 pairs of test images with different rotation angles and different scale transformations.

\subsubsection{Evaluation Metrics}
In order to quantitatively analyze the performance of our method, we use the number of correctly matched points (NCM), root mean square error (RMSE), and matching success rate (SR) defined in~\cite{li2023multimodal} as evaluation metrics. The NCM indicates that a point with residual difference less than a certain pixel under the truth transform is considered to be a correctly matched point pair, and the SR indicates that a correctly matched point pair is considered to be successfully matched if the number of correctly matched points satisfies $NCM > 10$, and the RMSE indicates the error of a correctly matched pair after the truth transform is computed, and if the match fails to be correctly matched if $NCM < 10$, then the RMSE is set to None, and will not be involved in the computation.

\subsection{Qualitative and quantitative comparison results }
\label{result}
\begin{figure}[h]
	\centering
	\includegraphics[width=\linewidth]{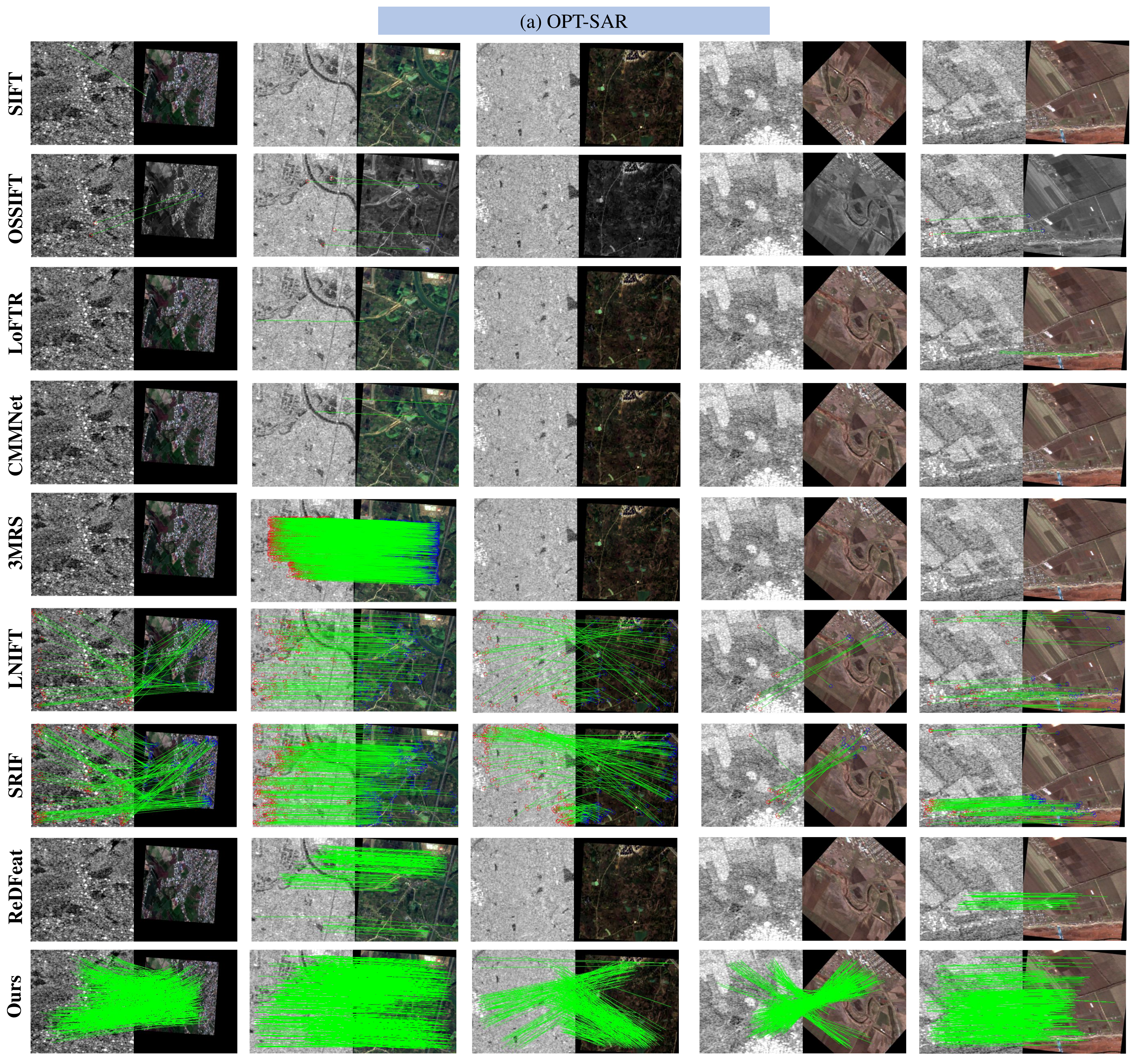}
	\caption{Results of qualitative comparisons of all tested methods for typical image pairs in the (a) OPT-SAR dataset. We show matched pairs with RMSE less than 3.}
	\label{sar1_img}
\end{figure}

\begin{figure}[h]
	\centering
	\includegraphics[width=\linewidth]{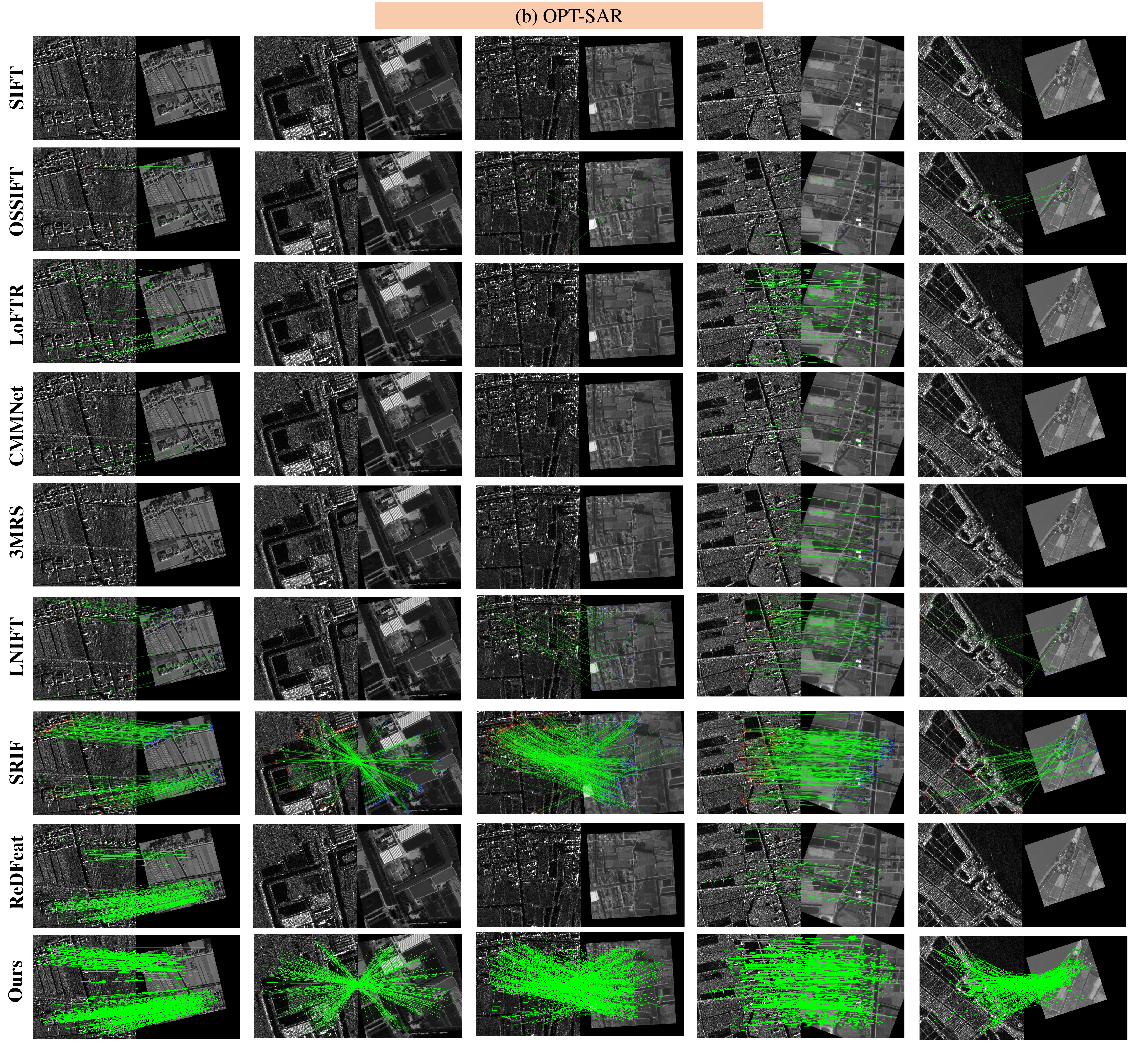}
	\caption{Results of qualitative comparisons of all tested methods for typical image pairs in the (b) OPT-SAR dataset. We show matched pairs with RMSE less than 3.}
	\label{sar2_img}
\end{figure}

\begin{figure}[h]
	\centering
	\includegraphics[width=\linewidth]{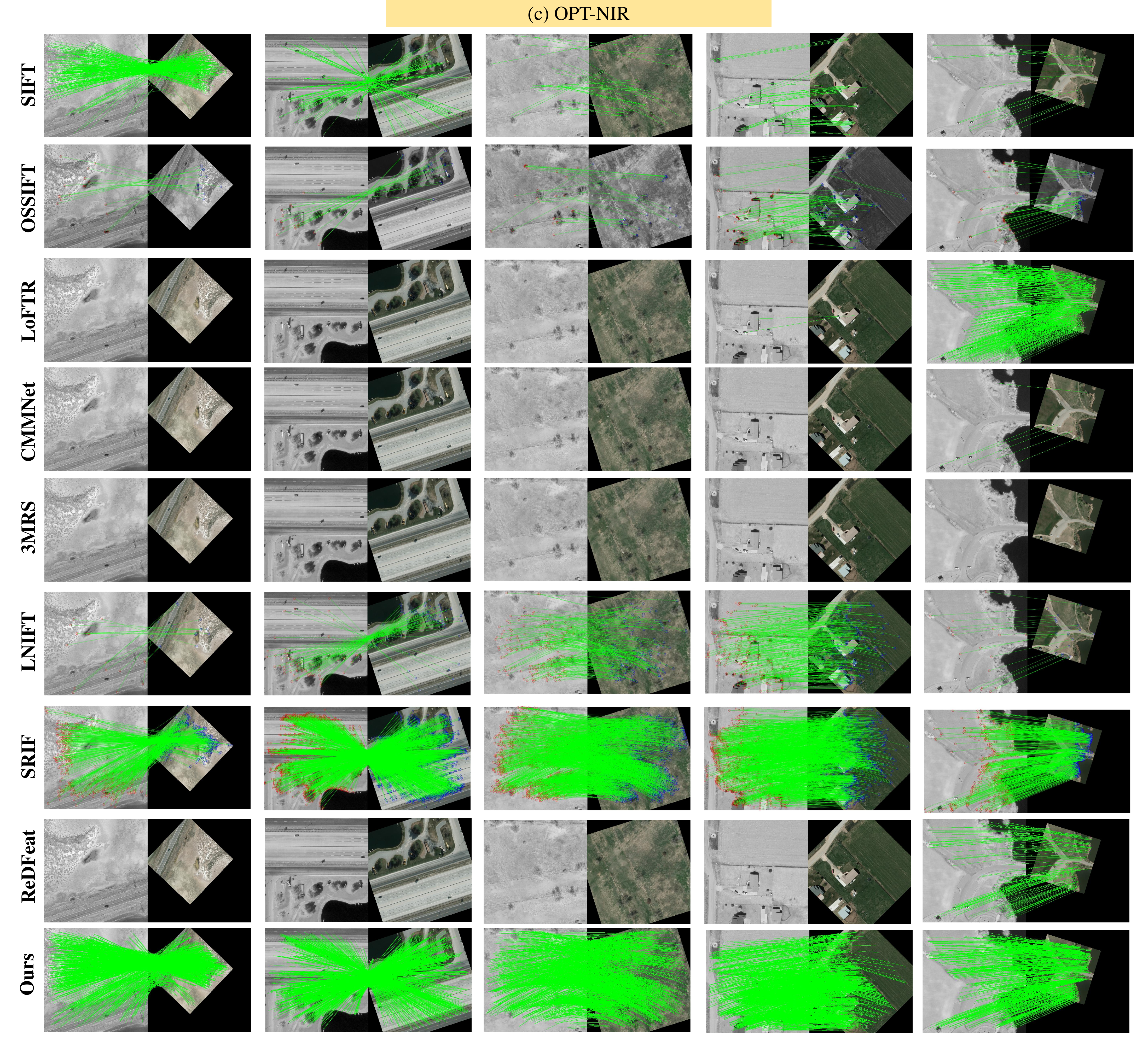}
	\caption{Results of qualitative comparisons of all tested methods for typical image pairs in the (c) OPT-NIR dataset. We show matched pairs with RMSE less than 3.}
	\label{nir1_img}
\end{figure}

\begin{figure}[h]
	\centering
	\includegraphics[width=0.95\linewidth]{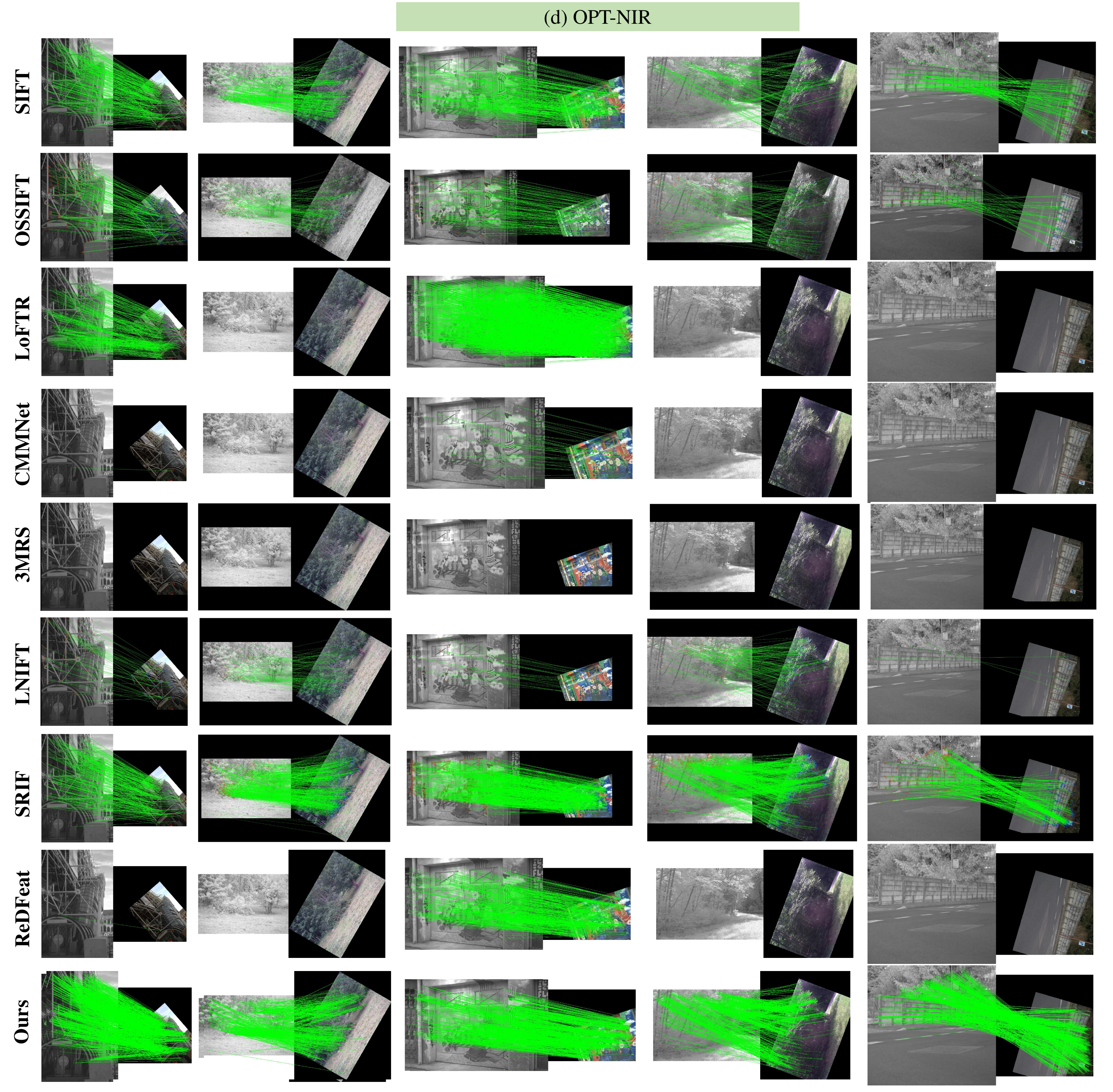}
	\caption{Results of qualitative comparisons of all tested methods for typical image pairs in the (d) OPT-NIR dataset. We show matched pairs with RMSE less than 3.}
	\label{nir2_img}
\end{figure}

We selected four independent datasets, resulting in a total of 33,180 pairs of test images. The modal and geometric differences between the optical and SAR images are relatively large, making accurate matching difficult, whereas the modal and geometric differences between the optical and NIR images are small, mainly presenting rotational and scale differences. We provide a visual comparison of typical image pairs from all test datasets in Fig.~\ref{sar1_img}~\ref{sar2_img}~\ref{nir1_img}~\ref{nir2_img}. In addition, to further analyze the performance of different methods for rotationally invariant matching in different angle intervals, we counted the average NCM, RMSE, and SR metrics by different angle intervals, as shown in Table ~\ref{tab:SAR1}~\ref{tab:SAR2}~\ref{tab:NIR1}~\ref{tab:NIR2}.

SIFT can achieve more stable results on the OPT-NIR dataset due to the small modal differences in NIR images and SIFT is robustness to rotation and scale. However, it is still affected by NRD, which leads to incorrect estimation of the main orientation, resulting in poor matching accuracy, as shown in Table ~\ref{tab:NIR1}~\ref{tab:NIR2}. In order to improve the matching accuracy of SIFT on optical and SAR images, OS-SIFT is proposed, and it can be seen that its performance is slightly better than that of SIFT on optical and SAR images, but it is still difficult to deal with the serious speckle noise, as shown in Table ~\ref{tab:SAR1}~\ref{tab:SAR2}. Meanwhile, it mainly targets optical and SAR images, and is difficult to deal with other multimodal images, and its performance is degraded compared with that of SIFT on the OPT-NIR dataset, as shown in Table ~\ref{tab:NIR1}~\ref{tab:NIR2}. LoFTR is a detector-free matching method that achieves superior results in weakly textured regions of homologous images. It has difficulty handling the modal differences and rotational scale differences in OPT-SAR images, but achieves better results with smaller rotations and scales when face OPT-NIR images with minor modal differences. CMMNet can also only achieve some results with small rotations. The 3MRS can achieve some results within the 30-degree interval, but has no matching effect in ranges greater than the 30-degree interval because it does not account for rotational and scale differences. As shown in the second pair of data in Fig.~\ref{sar1_img}, it achieves excellent results because 3MRS uses the dense template matching method to obtain more matched pairs of points, highlighting that the dense template matching method can significantly improve matching accuracy. The overall matching success rate of LNIFT exceeds 80\%, demonstrating superior results. The primary reason for the lower success rate in some instances may be its do not construct scale space, making it challenging to handle image pairs with scale transformations. Based on LNIFT, SRIF constructs a scale space and employs a local intensity binary transform (LIBT) to extract modal invariant features, which can improve the success rate (SR) of image matching from 84.7\% to 99.9\%. In contrast, our REMM method achieves an average success rate of 98.2\%, with the highest NCM and lowest RMSE. There are two main reasons for the superior performance of REMM: first, it achieves rotation and scale invariance through multimodal feature learning. This approach extracts more robust modal invariant features between multimodal images. Secondly, our cyclic shift module efficiently rotationally encodes descriptors while maintaining their discriminative properties. We also achieve competitive results when compared to current state-of-the-art methods.

The average success rates of the nine compared algorithms on the four datasets are 47.79\%, 48.28\%, 20.8\%, 4.84\%, 8.82\%, 84.76\%, 99.90\%, 16.59\%, and 98.19\%. Our method is slightly lower than SRIF mainly because data-driven learning methods are susceptible to the dataset. In terms of correctly matching points, all methods (SIFT, OS-SIFT, LoFTR, CMMNet, 3MRS, LNIFT, SRIF, ReDFeat, REMM) have average NCM values of 132, 72, 299, 6, 284, 105, 690, 160, and 1493, respectively. Our number of correctly matched points is approximately twice that of the sub-optimal result SRIF. In terms of RMSE of matched point pairs, the RMSE of all methods are 10.418, 5.619, 7.105, 15.483, 15.466, 2.007, 1.896, 11.612, and 1.550. Our method achieves the best RMSE results. Once again, the performance of our proposed method for rotation-equivariant matching is demonstrated.
\subsection{Performance of REMM on an independent dataset}
We demonstrate the effectiveness of our REMM method in comparative experiments on four datasets totaling 33,180 image pairs. For practical applications, researchers may be interested in the performance of our REMM method when applied to unknown datasets. In this section, we test our REMM using the independent test dataset provided in the paper~\cite{zhang2023histogram}, which consists of five types of data. Given our known training datasets are OPT-SAR and OPT-NIR, we select Depth-Optical images, SAR-Optical images, and Night-Day images as three types of multimodal images, testing a total of 30 image pairs.

\begin{table*}[!h]
	\caption{The quantitative comparison results on (a) OPT-SAR datasets. The top two results for NCM and SR are marked with red and blue.
		\label{tab:SAR1}}
	\centering
	\renewcommand{\arraystretch}{1.3}
	\footnotesize
	\scalebox{0.8}{
		\begin{tabular}{ccccccccccccc}
			\toprule[0.5pt]
			\multicolumn{11}{c}{\textit{(a)} OPT-SAR datasets}                                             \\ \hline
\multicolumn{1}{c}{Different angle ranges}&Metric  &SIFT &OS-SIFT& LoFTR& CMMNet & 3MRS& LNIFT &SRIF &ReDFeat &Ours\\\hline
			\multirow{3}{*}{\centering[-180\degree:-90\degree)}
&NCM$\uparrow$  &0&0&0&0&0&29&{\color{blue}123}&0&{\color{red}672} \\
&RMSE$\downarrow$  &20&20&20&20&20&2.105&2.120&20&1.874\\
&SR(\%)$\uparrow$  &0&0&0&0&0&70.5&{\color{red}99.6}&0&{\color{blue}96.2} \\\hline
			\multirow{3}{*}{\centering[-90\degree:-30\degree)}
&NCM$\uparrow$ &0&0&14&0&13&57&{\color{blue}149}&14&{\color{red}700} \\
&RMSE$\downarrow$  &20&20&1.753&20&2.077&2.108&2.120&1.824&1.871\\
&SR(\%)$\uparrow$  &0&0&0.3&0&0.1&96.5&{\color{red}99.5}&0.3&{\color{blue}95.9} \\\hline
			\multirow{3}{*}{\centering[-30\degree:30\degree)}
&NCM$\uparrow$  &0&11&19&0&340&93&{\color{blue}190}&86&{\color{red}749} \\
&RMSE$\downarrow$ &20&2.213&1.810&20&1.979&2.103&2.119&1.695&1.872\\
&SR(\%)$\uparrow$ &0&0.1&18.8&0&26.5&98.9&{\color{red}99.8}&46.2&{\color{blue}95.8} \\\hline
			\multirow{3}{*}{\centering[30\degree:90\degree)}
&NCM$\uparrow$  &0&0&14&0&0&57&{\color{blue}157}&0&{\color{red}711} \\
&RMSE$\downarrow$  &20&20&1.444&20&20&2.100&2.119&20&1.869\\
&SR(\%)$\uparrow$  &0&0&0.1&0&0&96.7&{\color{red}99.3}&0&{\color{blue}95.7} \\\hline
			\multirow{3}{*}{\centering[90\degree:180\degree]}
&NCM$\uparrow$  &0&0&0&0&0&28&{\color{blue}124}&0&{\color{red}665} \\
&RMSE$\downarrow$  &20&20&20&20&20&2.103&2.121&20&1.872\\
&SR(\%)$\uparrow$  &0&0&0&0&0&70.9&{\color{red}99.9}&0&{\color{blue}95.6} \\\hline
            \toprule[1.0pt]
	    \end{tabular}}
\end{table*}

\begin{table*}[!h]
	\caption{The quantitative comparison results on (b) OPT-SAR datasets. The top two results for NCM and SR are marked with red and blue.
		\label{tab:SAR2}}
	\centering
	\renewcommand{\arraystretch}{1.3}
	\footnotesize
	\scalebox{0.8}{
		\begin{tabular}{ccccccccccccc}
			\toprule[1pt]
			\multicolumn{11}{c}{\textit{(b)} OPT-SAR datasets}                                             \\ \hline
\multicolumn{1}{c}{Different angle ranges}&Metric  &SIFT &OS-SIFT& LoFTR& CMMNet & 3MRS& LNIFT &SRIF &ReDFeat &Ours\\\hline
			\multirow{3}{*}{\centering[-180\degree:-90\degree)}
&NCM$\uparrow$  &0&13&0&0&0&22&{\color{blue}174}&0&{\color{red}650} \\
&RMSE$\downarrow$  &20&2.035&20&20&20&2.09&2.067&20&1.897\\
&SR(\%)$\uparrow$  &0&1&0&0&0&44.8&{\color{red}100}&0&{\color{red}100} \\\hline
			\multirow{3}{*}{\centering[-90\degree:-30\degree)}
&NCM$\uparrow$  &0&14&37&0&0&38&{\color{blue}206}&0&{\color{red}682} \\
&RMSE$\downarrow$ &20&2.056&1.866&20&20&2.096&2.066&20&1.896\\
&SR(\%)$\uparrow$ &0&9&21.9&0&0&{\color{blue}87.5}&{\color{red}100}&0&{\color{red}100} \\\hline
			\multirow{3}{*}{\centering[-30\degree:30\degree)}
&NCM$\uparrow$  &0&14&40&14&{\color{red}1107}&50&231&91&{\color{blue}721} \\
&RMSE$\downarrow$  &20&2.088&1.850&1.941&1.976&2.092&2.066&1.836&1.894\\
&SR(\%)$\uparrow$  &0&22.1&55.5&8.4&41.6&{\color{blue}93.5}&{\color{red}100}&67.9&{\color{red}100} \\\hline
			\multirow{3}{*}{\centering[30\degree:90\degree)}
&NCM$\uparrow$  &0&13&25&0&0&36&{\color{blue}199}&0&{\color{red}688} \\
&RMSE$\downarrow$  &20&2.091&1.899&20&20&2.098&2.065&20&1.890\\
&SR(\%)$\uparrow$  &0&10.8&6.3&0&0&{\color{blue}87.3}&{\color{red}100}&0&{\color{red}100} \\\hline
			\multirow{3}{*}{\centering[90\degree:180\degree]}
&NCM$\uparrow$  &0&12&0&0&0&22&{\color{blue}179}&0&{\color{red}645} \\
&RMSE$\downarrow$  &20&2&20&20&20&2.093&2.067&20&1.899\\
&SR(\%)$\uparrow$  &0&10&0&0&0&{\color{blue}43.1}&{\color{red}100}&0&{\color{red}100} \\\hline
            \toprule[1.0pt]
	    \end{tabular}}
\end{table*}

\begin{table*}[!h]
	\caption{The quantitative comparison results on (c) OPT-NIR datasets. The top two results for NCM and SR are marked with red and blue.
		\label{tab:NIR1}}
	\centering
	\renewcommand{\arraystretch}{1.3}
	\footnotesize
	\scalebox{0.8}{
		\begin{tabular}{ccccccccccccc}
			\toprule[1pt]
			\multicolumn{11}{c}{\textit{(c)} OPT-NIR datasets}                                             \\ \hline
\multicolumn{1}{c}{Different angle ranges}&Metric  &SIFT &OS-SIFT& LoFTR& CMMNet & 3MRS& LNIFT &SRIF &ReDFeat &Ours\\\hline
			\multirow{3}{*}{\centering[-180\degree:-90\degree)}
&NCM$\uparrow$ &159&43&20&0&0&98&{\color{blue}1248}&0&{\color{red}1742} \\
&RMSE$\downarrow$  &0.695&2.145&1.743&20&20&1.887&1.585&20&0.881\\
&SR(\%)$\uparrow$  &{\color{blue}97.1}&79.1&0.4&0&0&88.2&{\color{red}100}&0&{\color{red}100} \\\hline
			\multirow{3}{*}{\centering[-90\degree:-30\degree)}
&NCM$\uparrow$  &159&106&842&14&0&228&{\color{blue}1416}&241&{\color{red}1787} \\
&RMSE$\downarrow$ &0.487&1.890&1.247&1.963&20&1.898&1.581&0.817&0.912\\
&SR(\%)$\uparrow$ &97.1&96.9&35.5&1.6&0&{\color{blue}98.3}&{\color{red}100}&8.2&{\color{red}100} \\\hline
			\multirow{3}{*}{\centering[-30\degree:30\degree)}
&NCM$\uparrow$  &160&138&{\color{blue}1979}&31&{\color{red}2124}&320&1563&1422&1873 \\
&RMSE$\downarrow$  &0.371&1.728&0.732&1.894&1.601&1.893&1.584&0.902&0.835\\
&SR(\%)$\uparrow$  &97.7&98.4&98.8&5.6&54.6&{\color{blue}99.6}&{\color{red}100}&94.4&{\color{red}100} \\\hline
			\multirow{3}{*}{\centering[30\degree:90\degree)}
&NCM$\uparrow$  &160&100&661&0&0&221&{\color{blue}1419}&55&{\color{red}1828} \\
&RMSE$\downarrow$  &0.511&1.923&1.319&20&20&1.895&1.584&0.853&0.866\\
&SR(\%)$\uparrow$  &97.4&96.2&29.3&0&0&{\color{blue}98.3}&{\color{red}100}&0.7&{\color{red}100} \\\hline
			\multirow{3}{*}{\centering[90\degree:180\degree]}
&NCM$\uparrow$ &158&39&36&0&0&99&{\color{blue}1249}&0&{\color{red}1742} \\
&RMSE$\downarrow$  &0.699&2.167&1.658&20&20&1.882&1.589&20&0.911\\
&SR(\%)$\uparrow$  &{\color{blue}97.4}&76.6&0.2&0&0&87.3&{\color{red}100}&0&{\color{red}100} \\\hline
            \toprule[1.0pt]
	    \end{tabular}}
\end{table*}

\begin{table*}[!h]
	\caption{The quantitative comparison results on (d) OPT-NIR datasets. The top two results for NCM and SR are marked with red and blue.
		\label{tab:NIR2}}
	\centering
	\renewcommand{\arraystretch}{1.3}
	\footnotesize
	\scalebox{0.8}{
		\begin{tabular}{ccccccccccccc}
			\toprule[1pt]
			\multicolumn{11}{c}{\textit{(d)} OPT-NIR datasets}                                             \\ \hline
\multicolumn{1}{c}{Different angle ranges}&Metric  &SIFT &OS-SIFT& LoFTR& CMMNet & 3MRS& LNIFT &SRIF &ReDFeat &Ours\\\hline
			\multirow{3}{*}{\centering[-180\degree:-90\degree)}
&NCM$\uparrow$ &370&80&0&0&0&74&{\color{blue}950}&0&{\color{red}2644} \\
&RMSE$\downarrow$  &1.188&2.121&20&20&20&1.955&1.817&20&1.595\\
&SR(\%)$\uparrow$  &94&80.7&0&0&0&80.2&{\color{red}100}&0&{\color{blue}98.5} \\\hline
			\multirow{3}{*}{\centering[-90\degree:-30\degree)}
&NCM$\uparrow$  &365&216&399&19&0&164&{\color{blue}1035}&214&{\color{red}2761} \\
&RMSE$\downarrow$ &1.082&1.986&1.697&1.925&20&1.870&1.812&1.406&1.550\\
&SR(\%)$\uparrow$ &93.8&99.4&19&5.4&0&91.1&{\color{red}100}&13.7&{\color{blue}98.6} \\\hline
			\multirow{3}{*}{\centering[-30\degree:30\degree)}
&NCM$\uparrow$  &373&315&1571&54&{\color{blue}2101}&226&1196&1063&{\color{red}3098} \\
&RMSE$\downarrow$  &1.048&1.882&1.370&1.932&1.684&1.963&1.810&1.481&1.511\\
&SR(\%)$\uparrow$  &93.8&99.3&{\color{red}100}&75.7&53.5&95.5&{\color{red}100}&{\color{blue}97.9}&96.7 \\\hline
			\multirow{3}{*}{\centering[30\degree:90\degree)}
&NCM$\uparrow$  &369&232&329&0&0&168&{\color{blue}1061}&17&{\color{red}2866} \\
&RMSE$\downarrow$  &1.101&1.964&1.713&20&20&1.961&1.814&1.430&1.535\\
&SR(\%)$\uparrow$  &93.8&100&29.9&0&0&89.9&{\color{red}100}&2.4&{\color{blue}95.4}\\\hline
			\multirow{3}{*}{\centering[90\degree:180\degree]}
&NCM$\uparrow$ &369&96&0&0&0&82&{\color{blue}942}&0&{\color{red}2647} \\
&RMSE$\downarrow$  &1.185&2.097&20&20&20&1.949&1.822&20&1.575\\
&SR(\%)$\uparrow$  &93.8&85.9&0&0&0&77.1&{\color{red}100}&0&{\color{blue}95.3} \\\hline
            \toprule[1.0pt]
	    \end{tabular}}
\end{table*}

\begin{table}[!h]
	\caption{Quantitative results of REMM on independent dataset~\cite{zhang2023histogram}. The top two results for NCM and SR are marked with red and blue. 
		\label{tab:howp}}
	\centering
	\renewcommand{\arraystretch}{1.3}
	\footnotesize
	\scalebox{0.8}{
		\begin{tabular}{ccccccc}
			\toprule[0.8pt]
\multicolumn{1}{c}{Type of image pair}&Metric &OS-SIFT & 3MRS& LNIFT &SRIF &Ours\\\hline
			\multirow{3}{*}{\centering Depth-Optical}
&NCM$\uparrow$ &47 &266 &47 &{\color{blue}319} &{\color{red}441}\\
&RMSE$\downarrow$ &1.993 &2.094 &2.054 &1.997 &1.883\\
&SR(\%)$\uparrow$ &100  &90 &70 &100 &{\color{red}100}\\\hline
			\multirow{3}{*}{\centering SAR-Optical}
&NCM$\uparrow$   &32 &{\color{red}478} &107 &287 &{\color{blue}356}\\
&RMSE$\downarrow$ &1.966 &2.100 &2.089 &2.059 &1.925\\
&SR(\%)$\uparrow$  &{\color{blue}70} &60 &47 &50 &{\color{red}100}\\\hline
			\multirow{3}{*}{\centering Night-Day}
&NCM$\uparrow$ &30 &{\color{red}546} &112 &{\color{blue}378} &244  \\
&RMSE$\downarrow$ &1.945 &2.083 &2.051 &2.064 &1.886 \\
&SR(\%)$\uparrow$ &80 &50 &80 &{\color{red}100} &{\color{blue}90}  \\\hline
            \toprule[1.0pt]
	    \end{tabular}}
\end{table}

\begin{figure}[h]
	\centering
	\includegraphics[width=\linewidth]{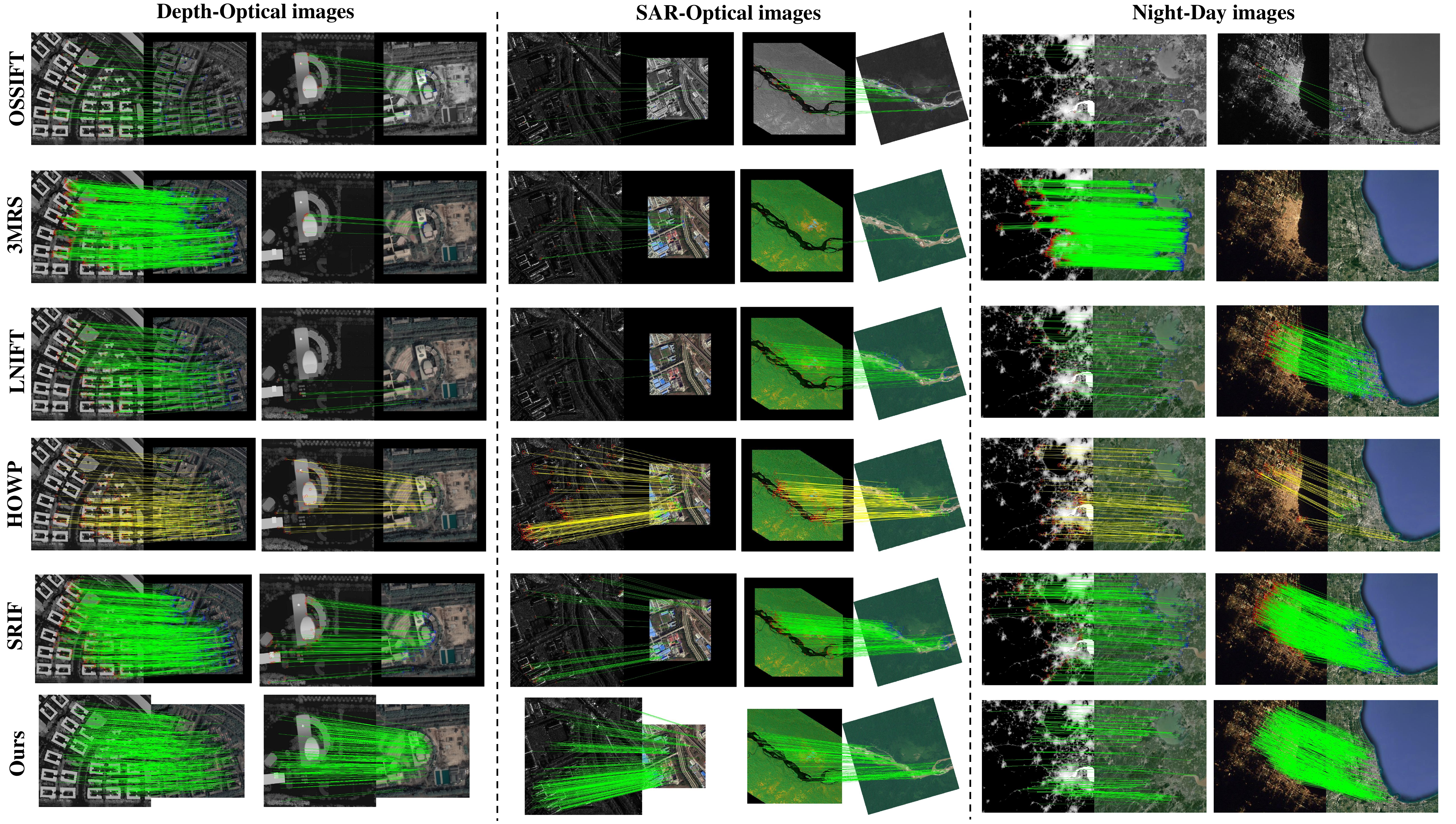}
	\caption{Results of qualitative comparisons of all tested methods for typical image pairs on the independent dataset~\cite{zhang2023histogram}. We show matched pairs with RMSE less than 3.}
	\label{INDATA}
\end{figure}

According to Table ~\ref{tab:howp} and Fig.~\ref{INDATA}, our approach demonstrates overall robustness. Our method REMM achieved 100\% matching success rate on Depth-Optical images and SAR-Optical images dataset, which outperforms current methods. However, it only achieved 90\% success rate in the Night-Day images dataset, due to the significant stylistic differences from the training data. Overall, our REMM achieved competitive results in both qualitative and quantitative comparison experiments, which once again demonstrates the effectiveness of our method in the multimodal image matching task.

\subsection{Ablation experiments}
\label{xr}
To assess the effectiveness of the crucial steps in the proposed REMM method, we performed ablation experiments on our benchmark test dataset and an independent dataset.

\subsubsection{Ablation study on group size of REMM}
In this section, we carry out an ablation study on our proposed benchmark to evaluate the effect of the group size parameter of our loop panning module, as shown in Table~\ref{tab:G}. We can see that the best results are achieved by dividing the 128-dimensional descriptors into 16 groups, each group includes 8-dimensional descriptors. the larger the group size is, the smaller the dimensionality of the descriptors included in each group, and the relative sensitivity of the finely shifted descriptors loses the rotational invariance of the descriptors, which in turn increases the difficulty of the matching. Therefore, when the group size is changed to 32 and 64, the effect decreases steadily. the smaller the group size is, the more descriptors are included in each group, and the rough translation makes it difficult to get rotational invariance of descriptors. So when group size is changed to 8, its effect also appears to decline.

\begin{table}[!h]
	\caption{Results of qualitative comparisons of Ablation result for Group size on ours benchmark. The top two results are marked with red and blue.  
		\label{tab:G}}
	\centering
	\renewcommand{\arraystretch}{1.3}
	\footnotesize
	\scalebox{0.8}{
		\begin{tabular}{ccccccc}
			\toprule[0.75pt]
\multicolumn{1}{c}{Image pair}&Group size &NCM$\uparrow$ &RMSE$\downarrow$ &SR(\%)$\uparrow$\\\hline
			\multirow{4}{*}{\centering Ours benchmark}
&G64 &1217 &1.625 &96.9\\
&G32 &{\color{blue}1310} &1.597 &97.7 \\
&G16 &{\color{red}1493} &{\color{red}1.550} &{\color{blue}98.2} \\
&G8 &1005  &{\color{blue}1.561} &{\color{red}99.0} \\
\hline
            \toprule[1.0pt]
	    \end{tabular}}
\end{table}

\begin{table*}[!h]
	\caption{Ablation result on TopK of REMM on independent dataset~\cite{zhang2023histogram}. The top two results for NCM and SR are marked with red and blue. 
		\label{tab:CFXR}}
	\centering
	\renewcommand{\arraystretch}{1.3}
	\footnotesize
	\scalebox{0.78}{
		\begin{tabular}{cccccccccccccc}
			\toprule[0.5pt]
\multicolumn{1}{c}{Type of image pair}&Metric &W/O cyclic shift  & TopK-1 &TopK-R0.9 &TopK-R0.8&TopK-R0.7&TopK-R0.6&TopK-R0.5&TopK-R0.4&TopK-R0.3&TopK-R0.2&TopK-R0.1\\\hline
			\multirow{3}{*}{\centering Depth-Optical}
&NCM$\uparrow$&348&180&214&256&298&334&370&394&414&{\color{blue}431}&{\color{red}441}\\
&RMSE$\downarrow$ &1.861&1.884&1.886&1.894&1.894&1.887&1.885&1.887&1.886&1.887&1.882\\
&SR(\%)$\uparrow$ &100&100&100&100&100&100&100&100&100&100&{\color{red}100}\\\hline
			\multirow{3}{*}{\centering SAR-Optical}
&NCM$\uparrow$   &213&127&160&197&230&268&294&319&335&{\color{blue}345}&{\color{red}356}\\
&RMSE$\downarrow$ &1.951&1.98&1.944&1.945&1.93&1.926&1.916&1.918&1.922&1.926&1.925\\
&SR(\%)$\uparrow$  &100&100&100&100&100&100&100&100&100&100&{\color{red}100}\\\hline
			\multirow{3}{*}{\centering Night-Day}
&NCM$\uparrow$ &127&77&107&126&145&166&186&205&219&{\color{blue}232}&{\color{red}243}  \\
&RMSE$\downarrow$ &1.881&1.873&1.875&1.877&1.892&1.903&1.902&1.889&1.896&1.897&1.885 \\
&SR(\%)$\uparrow$ &80&90&90&90&90&90&90&90&90&90&{\color{red}90}  \\\hline
			\multirow{1}{*}{\centering Average inference time}
&Time(s)$\downarrow$ &{\color{blue}0.734}&{\color{red}0.594}&0.85&0.943&1.091&1.277&1.55&1.917&2.427&3.051&3.926  \\\hline
            \toprule[1.0pt]
	    \end{tabular}}
\end{table*}

\subsubsection{Ablation study on TopK of REMM}
\begin{figure}[h]
	\centering
	\includegraphics[width=0.8\linewidth]{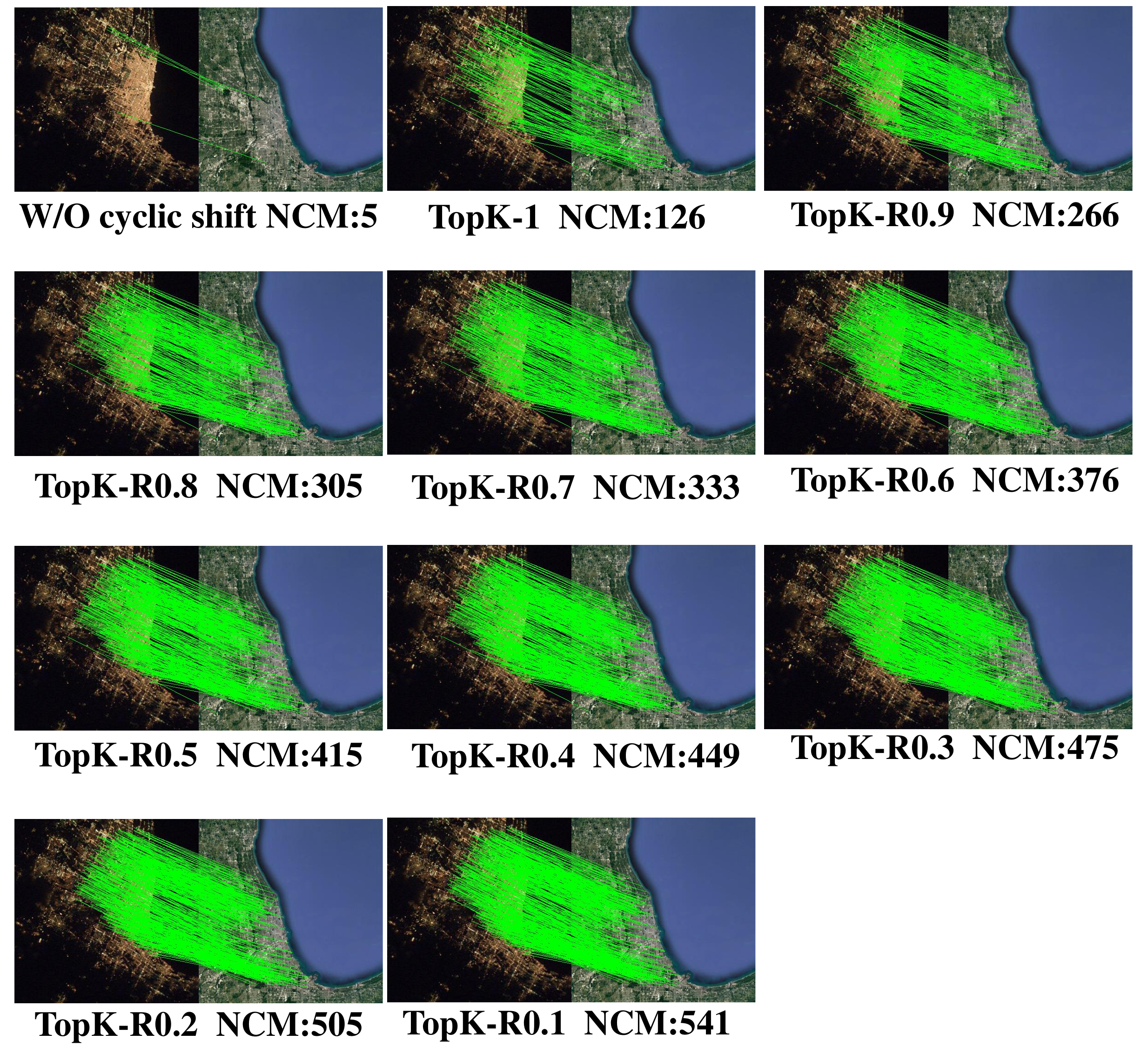}
	\caption{Results of qualitative comparisons of Ablation result on TopK on the independent dataset~\cite{zhang2023histogram}. We show matched pairs with RMSE less than 3.}
	\label{xr2_img}
\end{figure}

In this section, we conduct an ablation study on an independent dataset~\cite{zhang2023histogram} to evaluate the effect of the TopK parameter of our cyclic shift module. We demonstrate the module's effectiveness for rotation-equivariant matching during the training phase through the group size ablation study. The parameter settings of TopK also affect the final matching performance in the testing phase. We set different TopK parameters to obtain quantitative and qualitative matching results for various data types, as shown in Fig.~\ref{xr2_img} and Table~\ref{tab:CFXR}. We can see that as the TopK-R value decreases, the matching accuracy gradually improves, but it also leads to an increase in the inference time. This occurs because a lower R-value will result in each descriptor receiving more shift values, resulting in more descriptors, thereby increasing the likelihood of a successful match. The Night-Day images dataset shows the most significant decrease in effectiveness when there is no cyclic shift. This is because this part of the test data contains large rotational and scale transformations, and the lack of cyclic shift results in descriptors that do not have better rotational invariance, as shown in Fig.~\ref{xr2_img}. This demonstrates the effectiveness of the cyclic shift module for the construction of rotation-equivariant descriptors.

\subsection{Average running time}
As shown in Table~\ref{tab:time}, we give the average running time of our proposed method (REMM) and other traditional manual methods on an independent dataset of 30 pairs of images. All experiments were run on Windows 10 64 professional system, Intel(R) Core(TM) i7-11800H @ 2.30GHz, 32GB RAM, NVIDIA GTX 3080, Matlab 2021a.
\begin{table}[!h]
	\caption{The average running times of the several methods on independent dataset~\cite{zhang2023histogram}.  
		\label{tab:time}}
	\centering
	\renewcommand{\arraystretch}{1.3}
	\footnotesize
	\scalebox{0.8}{
		\begin{tabular}{ccccccc}
			\toprule[0.75pt]
\multicolumn{1}{c}{Methods}&OS-SIFT &3MRS &LNIFT &SRIF &Ours\\\hline
			\multirow{1}{*}{\centering Time (s)}
&10.043 &2.820  &0.782 &8.831 &3.926\\\hline
            \toprule[1.0pt]
	    \end{tabular}}
\end{table}
We can see that our proposed method REMM is comparable to 3MRS in performance, with our running time being second only to LNIFT. This is mainly due to the fact that REMM can obtain deep descriptors by combining DNNs without shared weights and DNN layers with shared weights, without requiring complex computations.

\section{Discusses}
\label{5}
In this section, we compare and analyze different approaches according to the design principles of different algorithms. We categorize them into traditional hand-crafted methods and learning-based methods.

Among the traditional hand-crafted methods, SIFT, as one of the classical algorithms for image matching, is based on gradient-based rotation and scale invariant descriptors. Its matching success rate is greater than 90\% only on (c) and (d) OPT-NIR data with small modal differences. It is almost invalid for (a) and (b) OPT-SAR data with large modal differences, primarily because the modal differences lead to unstable gradients, making it difficult for SIFT to achieve good results. Subsequently, OS-SIFT was proposed to match optical and SAR images, which improves the accuracy of gradient computation in SAR images and shows better results than SIFT on OPT-SAR data, but it is limited to optical and SAR images. 3MRS designed a dense template matching method to achieve good results for multimodal images without rotation and scale transformation, that is because rotation and scale transformations are not considered in the design. LNIFT reduces the modal differences by local normalization and adopts ORB main orientation estimation method to construct rotation-invariant descriptors, but it is susceptible to the influence of the still existing NRD, resulting in poor matching accuracy, and does not account for scale changes. Subsequently, SRIF utilized local intensity binary transform and scale space, significantly improving the success rate of matching, but its number of correct matching points is small, which is mainly limited by the robustness of the extracted features.

Among the learning-based methods, CMMNet uses deep learning in the field of multimodal image matching, which shows some results in the absence of rotation and scale transformations but struggles with complex modal differences and rotation-scale differences. LoFTR adopts a detector-free matching method, introducing the use of transformers to achieve dense matching results in weakly textured regions. It achieves excellent results (c) and (d) OPT-NIR data with small modal and rotational differences, with 98.8\% and 100\% matching success rates, respectively. The potential of the detector-free matching method for image matching is also demonstrated, but it struggles to obtain correct matching results when (a) and (b) OPT-SAR data modal or rotation differences are large. RedFeat proposed a keypoint detection and description coupling method for multimodal image matching, but it showed very high sensitivity to rotation. Considerable results can be achieved for all multimodal images without rotation, and matching completely fails in the presence of large rotations.

Our method draws on the strengths of both ReDfeat and traditional hand-crafted methods. Modal invariant features between multimodal images are learned through a multimodal feature learning module. We also introduce the cyclic shift module to rotationally encode the modal invariant descriptors, thereby obtaining rotation-equivariant descriptors, which makes them robust to any angle. To address scale transformations in multimodal images, we make the matching scale-invariant by constructing a scale space. These measures significantly improve the robustness and reliability of REMM, yielding encouraging results when compared with state-of-the-art methods. However, our proposed REMM method still faces challenges with large-scale transformations, and we will continue to investigate robust matching methods under large-scale transformations in the future.

\section{Conclusion}
\label{6}
In this paper, we propose a rotation-equivariant framework for end-to-end multimodal image matching, called REMM. Previous learning-based methods mainly focus on extracting modal-invariant descriptors, while consistently ignoring the rotational invariance. REMM learns the modal invariant features through a multimodal feature learning module and then employs a cyclic shift module to rotationally encode the modal invariant descriptors, thereby obtaining rotation-equivariant descriptors, which makes them robust to any angle. Extensive experiments and comprehensive analyses of our proposed method demonstrate the effectiveness of rotation-equivariant matching. Numerous experiments on our proposed benchmark and independent datasets demonstrate that REMM outperforms current multimodal image matching methods with strong generalization ability.

\bibliographystyle{IEEEtran}
\bibliography{main}
\end{document}